\documentclass{article}
 \usepackage[final]{nips_2016}
\usepackage{epstopdf}
\usepackage{float}

\usepackage[T1]{fontenc}    
\usepackage{hyperref}       
\usepackage{url}            
\usepackage{booktabs}       
\usepackage{amsfonts}       
\usepackage{nicefrac}       
\usepackage{microtype}      

\usepackage{qiangstyle}

\title{Bootstrap Model Aggregation for Distributed Statistical Learning}
\author{
  Jun Han \\
  Department of Computer Science\\
  Dartmouth College\\
  \texttt{jun.han.gr@dartmouth.edu} \\
   \And
   Qiang Liu \\
  Department of Computer Science\\
  Dartmouth College \\
   \texttt{qiang.liu@dartmouth.edu} \\
}
\begin{document}
\maketitle
\begin{abstract}
In distributed, or privacy-preserving learning, we are often given a set of probabilistic models estimated from different local repositories, and asked to combine them into a single model that gives efficient statistical estimation.
A simple method is to linearly average the parameters of the local models, which, however, tends to be degenerate or not applicable on non-convex models, or models with different parameter dimensions.
One more practical strategy is to generate bootstrap samples from the local models, and then learn a joint model based on the combined bootstrap set.
Unfortunately, the bootstrap procedure introduces additional noise and can significantly deteriorate the performance.
In this work, we propose two variance reduction methods to correct the bootstrap noise, including a weighted M-estimator that is both statistically efficient and practically powerful. Both theoretical and empirical analysis is provided to demonstrate our methods.
\end{abstract}

\section{Introduction}

\todo{"did not follow your logic flow", "it is fine to adopt sentence from other papers, but need to modify, especially this is in first sentence"
The Big-Data era, characterized by huge datasets and an appetite for new scientific and business insights, often involves learning statistical models of high complexity.
The rapid growth in the size and scale of datasets makes it impractical to perform these datasets in a single machine due to the speed and memory concerns \citep{mitra2011characterizing, balcan2012distributed}.
Meanwhile, in many contemporary applications, data are often partitioned across multiply servers \citep{corbett2013spanner}.
This attracts enormous attention to distributed statistical learning \citep{dekel2012optimal, duchi2012dual, huang2015distributed}.
 The advantage of distributed statistical inference is communication-efficient compared with traditional methods that transmit data to a central machine and then apply centralized algorithms. For example, the speed of local processors can be thousands time faster than the speed of data transmission in a modern network \citep{jaggi2014communication}. In some applications, data have to be stored in different locations because of privacy concerns.
 }
Modern data science applications increasingly involve learning complex probabilistic models over massive datasets.
In many cases, the datasets are distributed into multiple machines at different locations, between which communication is expensive or restricted; 
this can be either because the data volume is too large to store or process in a single machine, or due to privacy constraints
as these in healthcare or financial systems.
There has been a recent growing interest in developing
\emph{communication-efficient} algorithms for probabilistic learning with distributed datasets; see e.g., \citet{boyd2011distributed, zhang2012communication, dekel2012optimal, liu2014distributed, rosenblatt2014optimality} and reference therein.

This work focuses on a \emph{one-shot} approach for distributed learning, in which we first learn a set of local models from local machines, and then combine them in a fusion center to form a single model that integrates all the information in the local models.
This approach is highly efficient in both computation and communication costs,
but casts a challenge in designing statistically efficient combination strategies.
Many studies have been focused on a simple \emph{linear averaging} method that linearly averages the parameters of the local models \citep[e.g.,][]{zhang2012communication, zhang2013information, rosenblatt2014optimality}; although nearly optimal asymptotic error rates can be achieved, this simple method tends to degenerate in
practical scenarios for models with non-convex log-likelihood or non-identifiable parameters (such as latent variable models, and neural models),
and is not applicable at all for models with non-additive parameters (e.g., when the parameters have discrete or categorical values, or the parameter dimensions of the local models are different).

A better strategy that overcomes all these practical limitations of {linear averaging} is the \emph{KL-averaging} method \citep{liu2014distributed, merugu2003privacy},
which finds a model that minimizes the sum of Kullback-Leibler (KL) divergence to all the local models. 
In this way, we directly combine the models, instead of the parameters.
The exact \emph{KL-averaging} is not computationally tractable because of the intractability of calculating KL divergences;
a practical approach is to draw  (bootstrap) samples from the given local models, and then learn a combined model based on all the bootstrap data.
Unfortunately, the bootstrap noise can easily dominate in this approach and we need a very large bootstrap sample size to obtain accurate results.
In Section~\ref{sec:two}, we show that the MSE of the estimator obtained from the naive way is $O(N^{-1}+(dn)^{-1})$,
where $N$ is the total size of the observed data, and
$n$ is bootstrap sample size of each local model 
and $d$ is the number of machines. This means that to ensure a MSE of $O(N^{-1})$, which is guaranteed by the centralized method and the simple linear averaging, we need $d n \gtrsim N$;
this is unsatisfying since $N$ is usually very large by assumption. 

\todo{you do not need $  N>n\times d$ to make $O(N^{-1}+d^{-1}n^{-2})$, since when $N <= nd$, it is $O(N^{-1}  + (dn)^{-1}) = O(N^{-1})$}
In this work, we use variance reduction techniques to cancel out the bootstrap noise and get better KL-averaging estimates.
The difficulty of this task is first illustrated using a relatively straightforward control variates method,
which unfortunately suffers some of the practical drawback of the {linear averaging} method due to the use of a linear correction term.
We then propose a better method based on a weighted M-estimator, which inherits all the practical advantages of \emph{KL-averaging}.
On the theoretical part, we show that our methods give a MSE of $O(N^{-1} + (dn^2)^{-1})$, 
which significantly improves over the original bootstrap estimator.
Empirical studies are provided to verify our theoretical results and demonstrate the practical advantages of our methods. 

 This paper is organized as follows. Section~\ref{sec:background} introduces the background, and Section~\ref{sec:two} introduces our methods and analyze their theoretical properties.
 We present numerical results in Section~\ref{sec:empirical} and conclude the paper in Section~\ref{sec:conclusion}.
 Detailed proofs can be found in the appendix.

\myempty{
there are some complex probabilistic models where the linear average method and its modifications aforementioned are not applicable, such as latent variables models and probabilistic models characterized by discrete and nonadditive parameters \citep{blei2003latent, tipping1999mixtures}. The linear average method tends to degenerate in the presence of practical issues such as non-convexity and non-i.i.d. data partitions \citep{liu2014distributed}. To overcome these limitations, Liu and Ihler propose a $\KL$ divergence method which is to minimize the difference between the underlying true model and the learned models from the local machines. This idea has been widely applied to minimize the difference between deep generative models and deep inference networks \citep{rezende2014stochastic, rezende2015variational}.  If the probabilistic model is from the exponential family, the estimator obtained from $\KL$-divergence based method exactly equals the global MLE. \citet{liu2014distributed} have also proved that the $\KL$-divergence based method achieves minimal information loss on curved exponential family. More importantly, $\KL$-divergence method can be applied to any complex models where linear averaging methods cannot work. Despite the nice property of $\KL$-divergence based method, it introduces the challenging integral. The naive way to approximate the integration is via Monte Carlo bootstrap sampling. The MSE of the estimator obtained from the naive way is $O(N^{-1}+(dn)^{-1})$, where $n$ is bootstrap sample size of each local model(different local models draw the same size of bootstrap sample) and $d$ is the number of machines. This means that to ensure the MSE of this estimator $O(N^{-1})$, $n\times d=N$ is required, which is impractical.
}
\myempty{
The main contribution of this paper is to propose two more accurate estimators via control variates method and importance sampling. The MSEs of these two estimators are $O(N^{-1}+d^{-1}n^{-2})$ when $N>n\times d$, which has been verified theoretically and numerically. To get $O(N^{-1})$ MSE we just need a small bootstrap sample size to get high accuracy, compared with the $\KL-$naive way. This improvement of accuracy is important and make distributed learning on latent variables model practical and robust. A lot of numerical experiments have been done to verify our theoretical analysis both on toy examples and real datasets.
}
\myempty{
 This paper is organized as follows. Section~\ref{sec:background} introduces the distributed learning problem and the $\KL$-naive estimator.
 We propose our two more accurate estimators and analyze their theoretical properties in section~\ref{sec:two}, and further demonstrate them using empirical results in Section~\ref{sec:empirical}. The conclusion is drawn in Section~\ref{sec:conclusion}.
 Detailed proofs can be found in the appendix.
}

\section{Background and Problem Setting} \todo{$q$ is used for both dimension $R^p$ and probability $p(x)$}
\label{sec:background}
Suppose we have a dataset $X=\{\boldsymbol{x}_j, ~ j=1,2,...,N\}$ of size $N$, \emph{i.i.d.} drawn from a probabilistic model $p(\vx | \vv{\theta}^*)$ within a parametric family $\mathcal{P}=\{p(\boldsymbol{x}|\boldsymbol{\theta}):\boldsymbol{\theta}\in\Theta\}$; here $\boldsymbol{\theta}^*$ is the unknown true parameter that we want to estimate based on $X$.
In the distributed setting, the dataset $X$ is partitioned into $d$ disjoint subsets, $X=\bigcup_{k=1}^d X^k$, where $X^
k$ denotes the $k$-th subset which we assume is stored in a local machine.
For simplicity, we assume all the subsets have the same data size ($N/d$).
\todo{``sample set'' sounds weird}

The traditional maximum likelihood estimator (MLE) provides a natural way for estimating the true parameter $\boldsymbol{\theta}^*$
based on the whole dataset $X$,
\begin{align} \label{globalMLE}
\text{Global MLE:}\quad \boldsymbol{\hat{\theta}}_{\mathrm{mle}}=\argmax_{\boldsymbol{\theta}\in\Theta}\sum_{k=1}^d\sum_{j=1}^{N/d}\log p(\vx^k_j\mid\boldsymbol{\theta}),
\quad\text{where } X^k=\{\vx^k_j\}.
\end{align}
However, directly calculating the global MLE is challenging due to the distributed partition of the dataset.
Although distributed optimization algorithms exist \citep[e.g.,][]{boyd2011distributed, shamir2013communication},
they require iterative communication between the local machines and a fusion center,
which can be very time consuming in distributed settings, for which
the number of communication rounds
forms the main bottleneck (regardless of the amount of information communicated at each round).

We instead consider a simpler \emph{one-shot} approach that first learns a set of local models based on each subset, and then send them to a fusion center in which
they are combined into a global model that captures all the information. We assume each of the local models is estimated using a MLE based on subset $X^k$ from the $k$-th machine:  %
\todo{fix the indexes}
\begin{align}\label{equ:localmle}
\text{Local MLE:}\quad \boldsymbol{\hat{\theta}}_k=\argmax_{\boldsymbol{\theta}\in\Theta}\sum_{j=1}^{N/d}\log p(\boldsymbol{x}^k_j\mid\boldsymbol{\theta}),~~\text{where}~~ k\in [d]=\{1,2,\cdots,d\}.
\end{align}
The major problem is how to combine these local models into a global model.
The simplest way is to linearly average all local MLE parameters:
$$\text{Linear Average:}\quad \boldsymbol{\hat{\theta}}_{\mathrm{linear}}=\frac{1}{d}\sum_{k=1}^d\boldsymbol{\hat{\theta}}_k.$$
Comprehensive theoretical analysis has been done for $ \boldsymbol{\hat{\theta}}_{\mathrm{linear}}$ \citep[e.g.,][]{zhang2012communication, rosenblatt2014optimality}, which show that it has an asymptotic MLE of $\E||\boldsymbol{\hat{\theta}}_{\mathrm{linear}} - \vv{\theta}^* || = O(N^{-1})$.
In fact, it is equivalent to the global MLE $ \boldsymbol{\hat{\theta}}_{\mathrm{mle}}$ up to the first order $O(N^{-1})$, and
several improvements have been developed to improve the second order term \citep[e.g.,][]{zhang2012communication, huang2015distributed}.

Unfortunately,  the linear averaging method can easily break down in practice,
or is even not applicable when the underlying model is complex.
For example, it may work poorly when the likelihood has multiple modes,
or when there exist non-identifiable parameters for which different parameter values correspond to a same model (also known as the label-switching problem);
models of this kind include latent variable models and neural networks, and appear widely in machine learning. 
In addition, the linear averaging method is obviously not applicable when the local models have different numbers of parameters (e.g., Gaussian mixtures with unknown numbers of components),
or when the parameters are simply not additive (such as parameters with discrete or categorical values).
Further discussions on the practical limitaions of the linear averaging method can be found in \citet{liu2014distributed}.

All these problems of linear averaging can be well addressed by a
 \emph{KL-averaging} method which averages the model (instead of the parameters)
 by finding a geometric center of the local models in terms of KL divergence \citep{merugu2003privacy, liu2014distributed}.
Specifically, it finds a model $p( \vv x  ~|~ \vv{\theta}_{\mathrm{KL}}^*)$ where  $\vv{\theta}_{\mathrm{KL}}^*$  is obtained by 
 $\vv{\theta}_{\mathrm{KL}}^* = \argmin_{\vv\theta} \sum_{k=1}^d\KL(p(\boldsymbol{x}|\boldsymbol{\hat{\theta}}_k)\mid\mid p(\boldsymbol{x}|\boldsymbol{\theta}))$, which is equivalent to,
\begin{equation}
\label{KLdivmax}
\text{Exact KL Estimator: }\quad \boldsymbol{\theta}_{\KL}^*=\argmax_{\boldsymbol{\theta}\in\Theta} \bigg\{ \eta(\boldsymbol{\theta})
\equiv \sum_{k=1}^d\int p(\boldsymbol{x}\mid \boldsymbol{\hat{\theta}}_k)\log
p(\boldsymbol{x}\mid \boldsymbol{\theta})d\boldsymbol{x} \bigg\}.
\end{equation}
\citet{liu2014distributed} studied the theoretical properties of the KL-averaging method, and showed that
 it exactly recovers the global MLE, that is, $\boldsymbol{\theta}_{\KL}^*={\boldsymbol{\hat\theta}}_{\mathrm{mle}}$, when the distribution family is a full exponential family,
 and achieves an optimal asymptotic error rate (up to the second order) among all the possible combination methods of $\{\vv{\hat\theta}_k\}$.

Despite the attractive properties,
the exact KL-averaging is not computationally tractable except for very simple models.
\citet{liu2014distributed} suggested a naive \emph{bootstrap} method for approximation:
it draws \emph{parametric bootstrap} sample $\{\boldsymbol{\widetilde{x}}^k_j\}_{j=1}^{n}$ from each local model $p(\boldsymbol{x}|\boldsymbol{\hat{\theta}}_k)$, $k\in [d]$ and use it to approximate each integral in \eqref{KLdivmax}. 
The optimization in \eqref{KLdivmax} then reduces to a tractable one,
\begin{equation}
\label{KLdivmaxapprox}
\text{KL-Naive Estimator:} \quad\boldsymbol{\hat{\theta}}_{\mathrm{KL}}=  \argmax_{\boldsymbol{\theta}\in\Theta}  \bigg \{ \hat{\eta}(\boldsymbol{\theta}) \equiv    \frac{1}{n} \sum_{k=1}^d \sum_{j=1}^n \log p(\boldsymbol{\widetilde{x}}^k_j\mid \boldsymbol{\theta}) \bigg\}.
\end{equation}
Intuitively, we can treat each $\widetilde X_k = \{\boldsymbol{\widetilde{x}}^k_j\}_{j=1}^{n}$ as an approximation of the original subset $X^k  = \{\boldsymbol{{x}}^k_j\}_{j=1}^{N/d}$, and hence can be used to approximate the global MLE in \eqref{globalMLE}.

 Unfortunately, as we show in the sequel, the accuracy of  $\boldsymbol{\hat{\theta}}_{\KL}$ critically depends on the bootstrap sample size $n$, and one would need $n$ to be nearly as large as the original data size $N/d$ to make $\boldsymbol{\hat{\theta}}_{\mathrm{KL}}$ achieve the baseline asymptotic rate $O(N^{-1})$ that the simple linear averaging achieves; this is highly undesirably since $N$ is often assumed to be large in distributed learning settings.

\section{Main Results}
\label{sec:two}
We propose two variance reduction techniques for improving the KL-averaging estimates and discuss their theoretical and practical properties.
We start with a concrete analysis on the $\KL$-naive estimator $\boldsymbol{\hat{\theta}}_{\KL},$ which was missing in \citet{liu2014distributed}.  
\begin{ass}
\label{assump}
 1. $\log p(\boldsymbol{x}\mid\boldsymbol{\theta}),$
 $\frac{\partial\log p(\boldsymbol{x}\mid\boldsymbol{\theta})}{\partial\boldsymbol{\theta}},$
 and $\frac{\partial^2\log p(\boldsymbol{x}\mid\boldsymbol{\theta})}{\partial\boldsymbol{\theta}\partial\boldsymbol{\theta}^\top}$
 are continuous for $\forall \boldsymbol{x}\in\mathcal{X}$ and $\forall \boldsymbol{\theta}\in\Theta;$ 2. $\frac{\partial^2\log p(\boldsymbol{x}\mid\boldsymbol{\theta})}{\partial\boldsymbol{\theta}\partial\boldsymbol{\theta}^\top}$ is positive definite and $C_1\leq \|\frac{\partial^2\log p(\boldsymbol{x}\mid\boldsymbol{\theta})}{\partial\boldsymbol{\theta}\partial\boldsymbol{\theta}^\top}\|\leq C_2$ in a neighbor of $\boldsymbol{\theta}^*$ for $\forall x\in\mathcal{X}$, and $C_1$, $C_2$ are some positive constans.
\end{ass}
\begin{thm}
\label{thm1}
\label{naivemethod}
 Under Assumption \ref{assump},  ${\boldsymbol{\hat\theta}}_{\KL}$ is a consistent estimator of $\boldsymbol{\theta}_{\KL}^*$ as $n\to\infty$, and
 $$\mathbb{E}({\boldsymbol{\hat\theta}}_{\KL}-\boldsymbol{\theta}_{\KL}^*)=o(\frac{1}{dn}),\quad \mathbb{E}\|{\boldsymbol{\hat\theta}}_{\KL}-\boldsymbol{\theta}_{\KL}^*\|^2=O(\frac{1}{dn}),$$
 where $d$ is the number of machines and $n$ is the bootstrap sample size for each local model $p(\boldsymbol{x}\mid\boldsymbol{\hat{\theta}}_k)$.
\end{thm}
The proof is in Appendix A.
Because the MSE between the exact $\KL$ estimator $\boldsymbol{\theta}_{\KL}^*$ and the true parameter $\boldsymbol{\theta}^*$ is
$O(N^{-1})$
as shown in \citet{liu2014distributed}, the MSE between $\boldsymbol{\hat{\theta}}_{\KL}$ and the true parameter $\boldsymbol{\theta}^*$ is
\begin{equation}
\label{globalmse}
\mathbb{E}\|\boldsymbol{\hat{\theta}}_{\KL}-\boldsymbol{\theta}^*\|^2\approx\mathbb{E}\|\boldsymbol{\hat{\theta}}_{\KL}-\boldsymbol{\theta}^*_{\KL}\|^2+\mathbb{E}\|\boldsymbol{\theta}_{\KL}^*-\boldsymbol{\theta}^*\|^2=O(N^{-1}+(dn)^{-1}).
\end{equation}
To make the MSE between $\boldsymbol{\hat{\theta}}_{\KL}$ and $\boldsymbol{\theta}^*$ equal $O(N^{-1})$, as what is achieved by the simple linear averaging, we need draw $d n \gtrsim N$ bootstrap data points in total, which is undesirable since $N$ is often assumed to be very large by the assumption of distributed learning setting (one exception is when the data is distributed due to privacy constraint, in which case $N$ may be relatively small).

Therefore, it is a critical task to develop more accurate methods that can reduce the noise introduced by the bootstrap process.
In the sequel, we introduce two variance reduction techniques to achieve this goal.
One is based a (linear) control variates method that improves $\boldsymbol{\hat{\theta}}_{\KL}$ using a linear correction term,
and another is a \emph{multiplicative} control variates method that modifies the M-estimator in \eqref{KLdivmaxapprox} by assigning each bootstrap data point with a positive weight to cancel the noise.
We show that both method achieves a higher $O(N^{-1} + (dn^2)^{-1})$ rate under mild assumptions,
while the second method has more attractive practical advantages. 

\subsection{Control Variates Estimator}
\label{sec:control}
The control variates method is a technique for variance reduction on Monte Carlo estimation \citep[e.g.,][]{wilson1984variance}.
It introduces a set of correlated auxiliary random variables with known expectations or asymptotics (referred as the control variates), to balance the variation of the original estimator.
In our case, since each bootstrapped subsample $\widetilde X^k =\{\boldsymbol{\widetilde{x}}^k_j\}_{j=1}^n$ is know to be drawn from the local model $p(\boldsymbol{x} \mid\boldsymbol{\hat\theta}_k)$, we can construct a control variate by re-estimating the local model based on $\widetilde X^k$:
%
\begin{align}
\text{Bootstrapped Local MLE:}\quad\boldsymbol{\widetilde{\theta}}_k=\argmax_{\boldsymbol{\theta}\in\Theta}\sum_{j=1}^n\log p(\boldsymbol{\widetilde{x}}^k_j\mid\boldsymbol{\theta}),\quad \mathrm{for}~~ k\in [d],\label{tildethea}
\end{align}
where $\boldsymbol{\widetilde{\theta}}_k$ is known to converge to $\boldsymbol{\hat{\theta}}_k$ asymptotically.
This allows us to define the following control variates estimator:
%
\begin{equation}
\label{KLControl}
\text{KL-Control Estimator:}\quad \boldsymbol{\hat{\theta}}_{\KL-C}=\boldsymbol{\hat{\theta}}_{\KL}+\sum_{k=1}^d\boldsymbol{\mathfrak{B}}_k(\boldsymbol{\widetilde{\theta}}_k-\boldsymbol{\hat{\theta}}_k),
\end{equation}
where $\boldsymbol{\mathfrak{B}_k}$ is a matrix chosen to minimize the asymptotic variance of $ \boldsymbol{\hat{\theta}}_{\KL-C}$;
our derivation shows that the asymptotically optimal $\boldsymbol{\mathfrak{B}_k}$ has a form of
\begin{equation}
\label{scorecoeff}
\boldsymbol{\mathfrak{B}}_k=-(\sum_{k=1}^dI(\boldsymbol{\hat{\theta}}_k))^{-1}I(\boldsymbol{\hat{\theta}}_k), \quad k\in [d],
\end{equation}
where $I(\boldsymbol{\hat{\theta}}_k)$ is the empirical Fisher information matrix of the local model $p(\vv x \mid \boldsymbol{\hat{\theta}}_k)$.
Note that this differentiates our method  from the typical control variates methods where $\boldsymbol{\mathfrak{B}}_k$ is instead estimated using empirical covariance between the control variates and the original estimator (in our case, we can not directly estimate the covariance because $\boldsymbol{\hat{\theta}}_{\KL}$ and  $\boldsymbol{\widetilde{\theta}}_k$ are not averages of i.i.d. samples).
The procedure of our method is summarized in Algorithm \ref{alg:kl-control}.
Note that the form of \eqref{KLControl} shares some similarity with the one-step estimator in \citet{huang2015distributed}, but
\citet{huang2015distributed} focuses on improving the linear averaging estimator, and is different from our setting.

We analyze the asymptotic property of the estimator $\boldsymbol{\hat{\theta}}_{\KL-C}$,  and summarize it as follows.
\begin{thm}
\label{Control}
 Under Assumption (\ref{assump}), $\boldsymbol{\hat{\theta}}_{\KL-C}$ is a consistent estimator of $\boldsymbol{\theta}_{\KL}^*$ as $n\to\infty,$
 and its asymptotic MSE is guaranteed to be smaller than the KL-naive estimator ${\boldsymbol{\hat\theta}}_{\KL}$, that is,
 $$
  n \mathbb{E}\|\boldsymbol{\hat{\theta}}_{\KL-C}-\boldsymbol{\theta}_{\KL}^*\|^2 < n  \mathbb{E}\|{\boldsymbol{\hat\theta}}_{\KL}-\boldsymbol{\theta}_{\KL}^*\|^2, ~~~~~~ \text{as}~~ n\to \infty.
  $$
  In addition, when $N> n\times d$,  the $\boldsymbol{\hat{\theta}}_{\KL-C}$ has \emph{``zero-variance''} in that
   $\mathbb{E}\|{\boldsymbol{\hat\theta}}_{\KL}-\boldsymbol{\theta}_{\KL}^*\|^2=O((dn^2)^{-1})$.
   Further, in terms of estimating the true parameter, we have
\begin{equation}
\label{globalklc}
\mathbb{E}\|\boldsymbol{\hat{\theta}}_{\KL-C}-\boldsymbol{\theta}^*\|^2=O(N^{-1}+(dn^{2})^{-1}).
\end{equation}
 \end{thm}
The proof is in Appendix B.
From \eqref{globalklc}, we can see that the MSE between $\boldsymbol{\hat{\theta}}_{\KL-C}$ and $\boldsymbol{\theta}^*$ reduces to $O(N^{-1})$
as long as $n \gtrsim (N/d)^{1/2}$, which is a significant improvement over the KL-naive method which requires $n  \gtrsim N/d$.
When the goal is to achieve an $O(\epsilon)$ MSE, we would just need to take $n \gtrsim 1/(d\epsilon)^{1/2}$ when $N > 1/\epsilon$, that is,
$n$ does not need to increase with $N$ when $N$ is very large.

 Meanwhile, because $\boldsymbol{\hat{\theta}}_{\mathrm{KL}-C}$ requires a linear combination of $\vv{\hat \theta}_k$,
 $\vv{\widetilde \theta}_k$ and $\vv{\hat \theta}_{\mathrm{KL}}$, it carries the practical drawbacks of the linear averaging estimator as we discuss in Section~\ref{sec:background}.
 %
This motivates us to develop another \emph{KL-weighted} method shown in the next section, which achieves the same asymptotical efficiency as $\boldsymbol{\hat{\theta}}_{\mathrm{KL}-C}$, while still
inherits all the practical advantages of \emph{KL-averaging}.
%
\begin{algorithm}[tb]
\caption{KL-Control Variates Method for Combining Local Models}
\label{alg:kl-control}
\begin{algorithmic}[1]
\STATE {\bfseries Input:} Local model parameters $\{\boldsymbol{\hat{\theta}}_k\}_{k=1}^d$.
\STATE {Generate bootstrap data $\{\boldsymbol{\widetilde{x}}^k_j\}_{j=1}^n$ from each $p(\boldsymbol{x}|\boldsymbol{\hat{\theta}}_k)$, for $k\in [d]$.}
\STATE{Calculate the KL-Naive estimator, $\boldsymbol{\hat{\theta}}_{\KL}=\argmax_{\boldsymbol{\theta}\in\Theta}\sum_{k=1}^d\frac{1}{n}\sum_{j=1}^n \log p(\boldsymbol{\widetilde{x}}^k_j| \boldsymbol{\theta}).$}
\STATE {Re-estimate the local parameters $\widetilde{\boldsymbol{\theta}}_k$ via \eqref{tildethea} based on the bootstrapped data subset $\{\boldsymbol{\widetilde{x}}^k_j\}_{j=1}^n$}, ~ for $k \in [d]$.
\STATE{Estimate the empirical Fish information matrix $I(\boldsymbol{\hat{\theta}}_k)=\frac{1}{n}\sum_{j=1}^n \frac{\partial{\log p(\boldsymbol{\widetilde{x}}_j^k|\boldsymbol{\hat{\theta}}_k)}}{\partial{\boldsymbol{\theta}}}{\frac{\partial{\log p(\boldsymbol{\widetilde{x}}_j^k|\boldsymbol{\hat{\theta}}_k)}}{\partial{\boldsymbol{\theta}}}}^\top$, for $k\in [d]$.}
\STATE{\bfseries Ouput:}
The parameter $\boldsymbol{\hat{\theta}}_{\KL-C}$ of the combined model is given by \eqref{KLControl} and \eqref{scorecoeff}.
\end{algorithmic}
\end{algorithm}

\subsection{KL-Weighted Estimator}
\label{sec:kl-weighted}
%
Our {KL-weighted} estimator is based on directly modifying the M-estimator for $\boldsymbol{\hat{\theta}}_{\mathrm{KL}}$ in \eqref{KLdivmaxapprox},
by assigning each bootstrap data point $\vv{\widetilde{x}}_j^k$ a positive weight
according to the probability ratio $p(\vv{\widetilde{x}}_j^k  \mid  \vv{\hat{\theta}}_k ) / p(\vv{\widetilde{x}}_j^k \mid \vv{\widetilde{\theta}}_k)$ of the actual local model $p(x | \vv{\hat{\theta}}_k)$ and the re-estimated model $p(x |\vv{\widetilde{\theta}}_k)$ in \eqref{tildethea}.
Here the probability ratio acts like a \emph{multiplicative} control variate \citep{nelson1987control}, which has the advantage of being positive and applicable to non-identifiable, non-additive parameters. Our estimator is defined as
%
\begin{equation}
\label{KLweigthed}
\text{KL-Weighted Estimator:}\quad
{\boldsymbol{\hat\theta}}_{\KL-W} =
\argmax_{\boldsymbol{\theta}\in\Theta} \bigg\{ \widetilde{\eta}(\boldsymbol{\theta})  \equiv \sum_{k=1}^d\frac1n\sum_{j=1}^n
\frac{p(\boldsymbol{\widetilde{x}}_j^k|\boldsymbol{\hat{\theta}}_k)}{p(\boldsymbol{\widetilde{x}}_j^k| \boldsymbol{\widetilde{\theta}}_k)}\log p(\boldsymbol{\widetilde{x}}_j^k|\boldsymbol{\theta}) \bigg\}.
\end{equation}
We first show that this weighted estimator $\widetilde{\eta}(\boldsymbol{\theta})$ gives a more accurate estimation of $\eta(\boldsymbol{\theta})$ in \eqref{KLdivmax} than the straightforward estimator $\hat{\eta}(\boldsymbol{\theta})$ defined in \eqref{KLdivmaxapprox} for any $\boldsymbol{\theta}\in\Theta$. 
\begin{lem}
As $n\to\infty$, $\widetilde{\eta}(\boldsymbol{\theta})$ is a more accurate estimator of $\eta(\boldsymbol{\theta})$ than $\hat{\eta}(\boldsymbol{\theta})$, in that
\begin{equation}
n \mathrm{Var}(\widetilde{\eta}(\boldsymbol{\theta}))\leq  n \mathrm{Var}(\hat{\eta}(\boldsymbol{\theta})), ~~~~~\text{as }n\to\infty,~~
\quad \text{for any } \boldsymbol{\theta}\in\Theta.
\end{equation}
\end{lem}\todo{Does the same thing holds for MSE instead of Variance?}
\todo{does the inequality holds strictly as that in Theorem 3.3???}
This estimator is motivated by \citet{henmi2007importance} in which the same idea is applied to reduce the asymptotic variance in importance sampling. 
Similar result is also found in \citet{hirano2003efficient}, in which  it is shown that a similar weighted estimator with estimated propensity score is more efficient than the estimator using true propensity score in estimating the average treatment effects. 
Although being a very powerful tool, results of this type seem to be not widely known in machine learning, except several applications in semi-supervised learning \citep{sokolovska2008asymptotics, kawakita2013semi}, and off-policy learning \citep{li2015toward}.

We go a step further to analyze the asymptotic property of our weighted M-estimator $\boldsymbol{\hat{\theta}}_{\KL-W}$ that maximizes $\widetilde{\eta}(\boldsymbol{\theta})$. It is natural to expect that the asymptotic variance of $\boldsymbol{\hat{\theta}}_{\KL-W}$ is smaller than that of $\boldsymbol{\hat{\theta}}_{\KL}$ based on maximizing $\hat{\eta}(\boldsymbol{\theta})$; this is shown in the following theorem. 
\begin{thm}
\label{thm3}
Under Assumption~\ref{assump}, $\boldsymbol{\hat{\theta}}_{\KL-W}$ is a consistent estimator of $\boldsymbol{\theta}_{\KL}^*$ as $n\to\infty,$ and has a better asymptotic variance than $\boldsymbol{\hat{\theta}}_{\KL}$, that is, 
$$
 n \mathbb{E}\|\boldsymbol{\hat{\theta}}_{\KL-W}-\boldsymbol{\theta}_{\KL}^*\|^2  \le  n \mathbb{E}\|\boldsymbol{\hat{\theta}}_{\KL}-\boldsymbol{\theta}_{\KL}^*\|^2,
~~~~~ \text{when $n \to \infty$}.
$$
When $N>n\times d$, we have $\mathbb{E}\|\boldsymbol{\hat{\theta}}_{\KL-W}-\boldsymbol{\theta}_{\KL}^*\|^2=O(({dn^2})^{-1})$ as $n\to\infty.$
Further, its MSE for estimating the true parameter $\vv{\theta}^*$ is
\begin{align}
\label{globalklw}
\mathbb{E}\|\boldsymbol{\hat{\theta}}_{\KL-W}-\boldsymbol{\theta}^*\|^2
=O(N^{-1}+(dn^2)^{-1}).
\end{align}
\end{thm}
The proof is in Appendix C.
This result is parallel to Theorem~\ref{Control} for the linear control variates estimator $\boldsymbol{\hat{\theta}}_{\KL-C}$.
 Similarly, it reduces to an $O(N^{-1})$ rate once we take $n \gtrsim  (N/d)^{1/2}$.
\todo{can these two estimators  be shown to be asymptotically equivalent? confirm?}

Meanwhile,  unlike the linear control variates estimator, 
$\boldsymbol{\hat{\theta}}_{\KL-W}$ inherits all the practical advantages of KL-averaging:
it can be applied whenever the KL-naive estimator can be applied, including for models with non-identifiable parameters, or with different numbers of parameters. The implementation of $\boldsymbol{\hat{\theta}}_{\KL-W}$ is also much more convenient (see Algorithm~\ref{alg:kl-weighted}), since it does not need to calculate the Fisher information matrix as required by Algorithm~\ref{alg:kl-control}.   \todo{Please use ref{label} for Algorithm next time}
\begin{algorithm}[tb]
\caption{KL-Weighted Method for Combining Local Models}
\label{alg:kl-weighted}
\begin{algorithmic}[1]
\STATE {\bfseries Input:} Local MLEs $\{\boldsymbol{\hat{\theta}}_k\}_{k=1}^d$.
\STATE {Generate bootstrap sample $\{\boldsymbol{\widetilde{x}}^k_j\}_{j=1}^n$ from each $p(\boldsymbol{x}|\boldsymbol{\hat{\theta}}_k)$, for $k\in[d].$}
\STATE {Re-estimate the local model parameter $\boldsymbol{\widetilde{\theta}}_k$ in \eqref{tildethea} based on bootstrap subsample $\{\boldsymbol{\widetilde{x}}^k_j\}_{j=1}^n$, for each $k\in[d].$}
\STATE {\bfseries Output:}  The parameter $\boldsymbol{\hat{\theta}}_{\KL-W}$ of the combined model is given by \eqref{KLweigthed}.
\end{algorithmic}
\end{algorithm}

\section{Empirical Experiments}\label{sec:empirical}
We study the empirical performance of our methods on both simulated and real world datasets. 
We first numerically verify the convergence rates predicted by our theoretical results using simulated data,
and then demonstrate the effectiveness of our methods in a challenging setting when the number of parameters of the local models are different as decided by Bayesian information criterion (BIC).
Finally, we conclude our experiments by testing our methods on a set of real world datasets.

The models we tested include
 probabilistic principal components analysis (PPCA), mixture of PPCA and Gaussian Mixtures Models (GMM).
 GMM is given by
 $p(\boldsymbol{x}\mid \boldsymbol{\theta})=\sum_{s=1}^m\alpha_s\mathcal{N}(\boldsymbol{\mu}_s, \Sigma_s)$ where $\boldsymbol{\theta}=(\alpha_s, \boldsymbol{\mu}_s, \Sigma_s).$
PPCA model is  defined with the help of a hidden variable $\vv t$,  $p(\vv x  ~| ~ \vv \theta) = \int p(\boldsymbol{x} ~|~ \boldsymbol{t}; ~  \vv\theta)  p(\vv t ~|~  \vv \theta) d \vv t$, 
where
$p(\boldsymbol{x}\mid\boldsymbol{t};~ \vv\theta)=\mathcal{N}( \vv x;  ~ \boldsymbol{\mu}+W\boldsymbol{t},\sigma^2),$ and $
p (\boldsymbol{t} \mid \vv\theta) =  \mathcal{N}(\vv t; ~ \boldsymbol{0}, \boldsymbol{I})$ and $\vv \theta = \{\boldsymbol{\mu}, ~W, ~\sigma^2\}$.
The mixture of PPCA is $p(\vv x \mid \vv\theta) = \sum_{s=1}^m\alpha_s p_s(\vv x \mid \vv\theta_s)$, where $\vv\theta = \{\alpha_s, \vv\theta_s\}_{s=1}^m$ and each $p_s(\vv x \mid \vv\theta_s)$ is a PPCA model.

Because all these models are latent variable models with unidentifiable parameters, the direct linear averaging method are not applicable.
For GMM, it is still possible to use a \emph{matched linear averaging} which matches the mixture components of the different local models by minimizing a symmetric $\KL$ divergence; the same idea can be used on our linear control variates method to make it applicable to GMM. On the other hand, because the parameters of PPCA-based models are unidentifiable up to arbitrary orthonormal transforms, linear averaging and linear control variates can no longer be applied easily.
We use expectation maximization (EM) to learn the parameters in all these three models. 


\subsection{Numerical Verification of the Convergence Rates}
\begin{figure}[h]
\begin{centering}
\begin{tabular}{ccccc}
\includegraphics[height=0.24\textwidth]{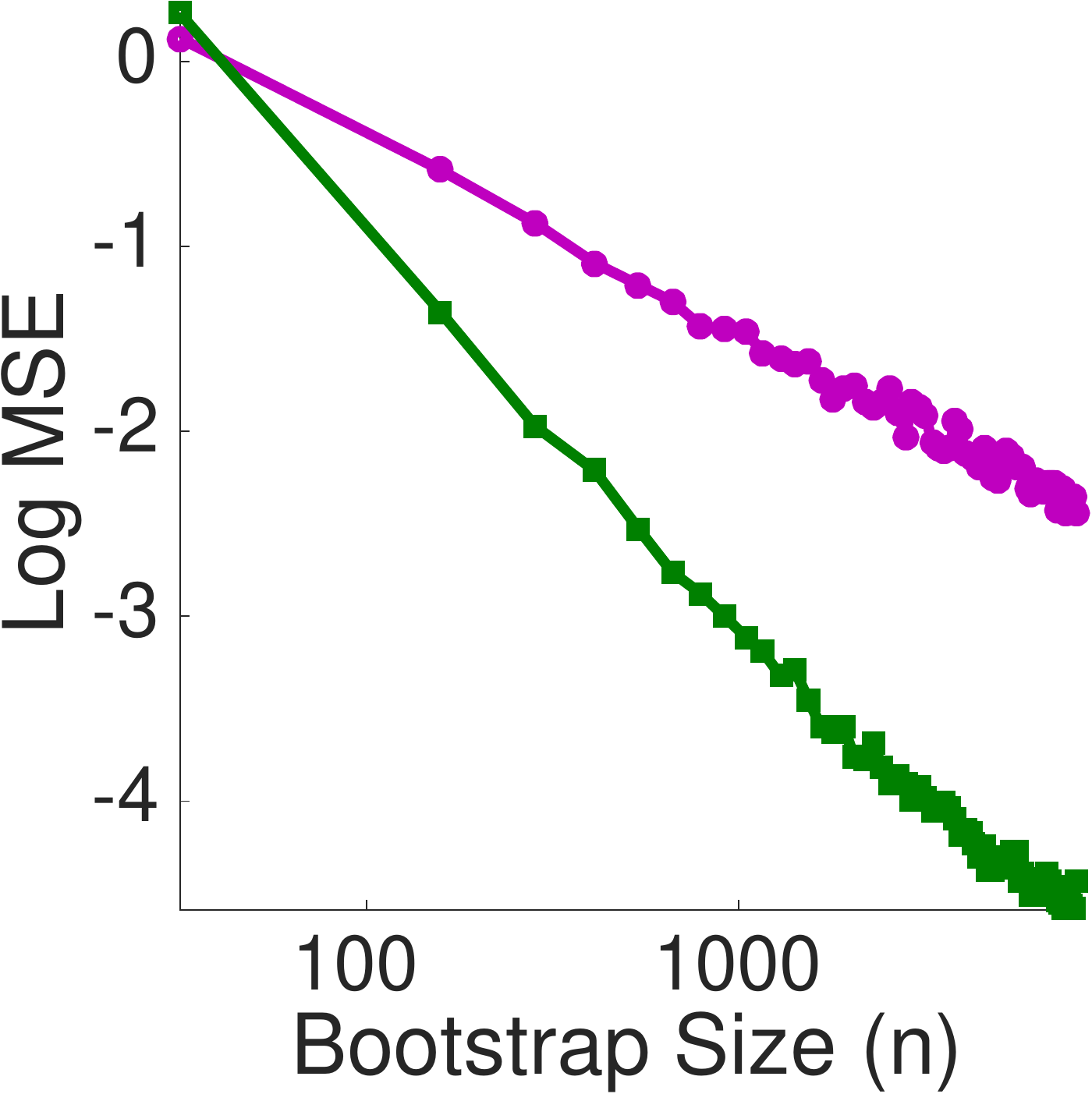}&&
\includegraphics[height=0.24\textwidth]{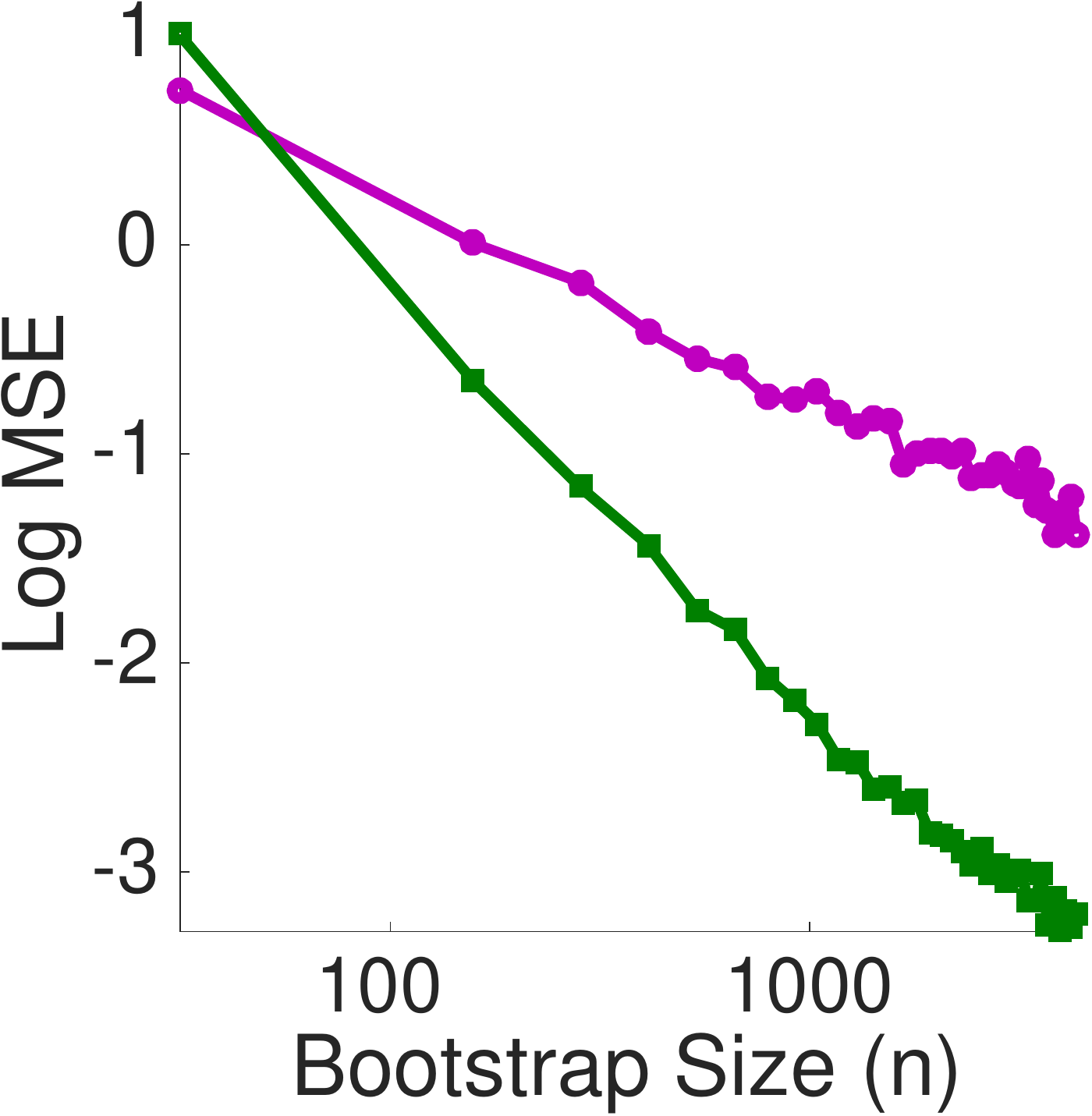}&&
\includegraphics[height=0.24\textwidth]{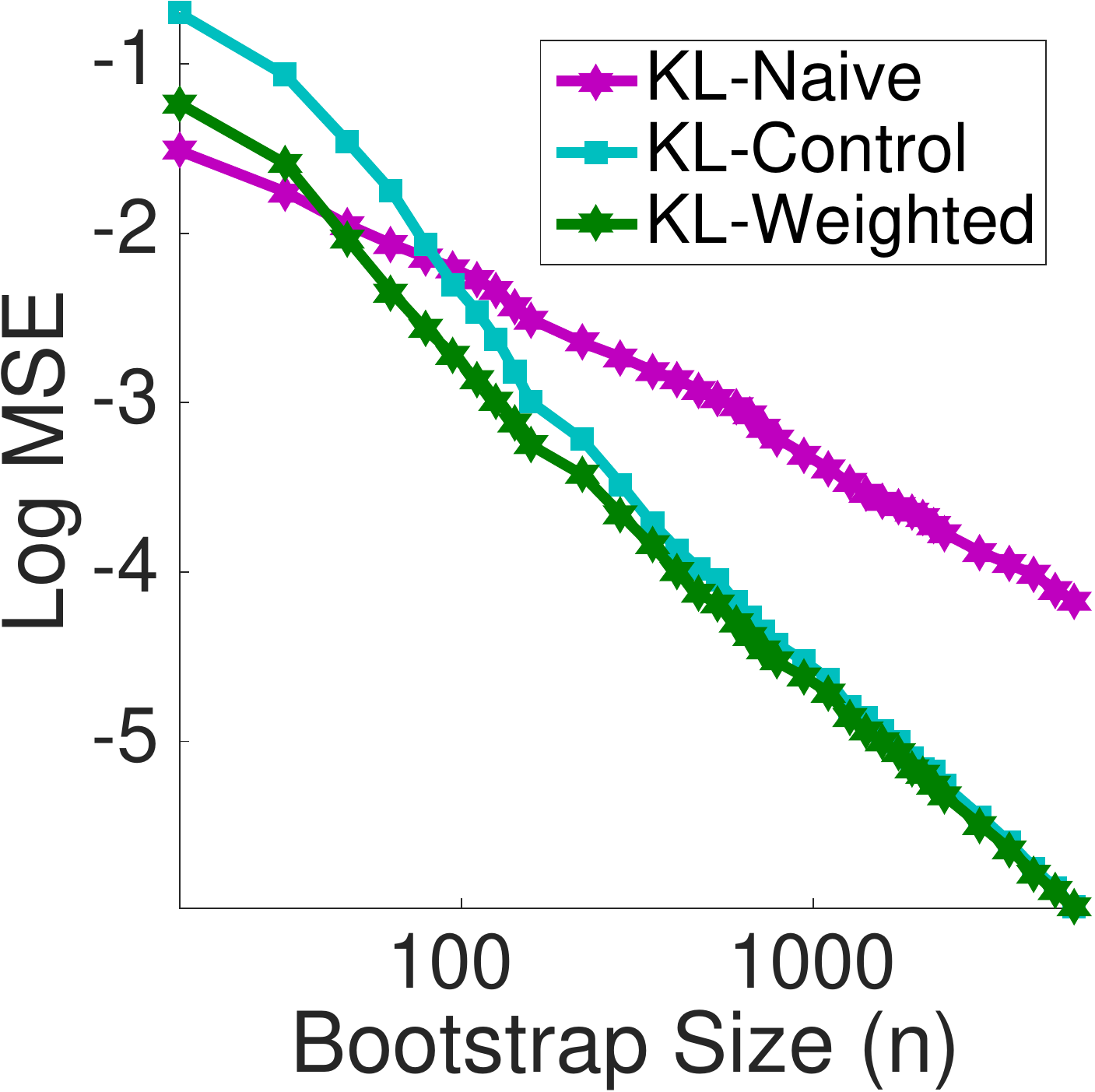} \\
{\small (a) PPCA} &&
{\small (b) Mixture of PPCA}&&
{\small (c) GMM }
\end{tabular}
\caption{
Results on different models with simulated data when we change the bootstrap sample size $n$, with fixed $d=10$ and $N=6\times10^7$. 
%
The dimensions of the PPCA models in (a)-(b) are 5, and that of GMM in (c) is 3.
The numbers of mixture components in (b)-(c) are 3.
Linear averaging and KL-Control are not applicable for the PPCA-based models, and are not shown in (a) and (b).
}
\label{fig:simple}
\end{centering}
\end{figure}

 We start with verifying the convergence rates
 in \eqref{globalmse}, \eqref{globalklc} and \eqref{globalklw}
 of MSE $\E||\vv{\hat\theta} -\vv \theta^*||^2$ of
 the different estimators for estimating the true parameters.
 Because there is also an non-identifiability problem in calculating the MSE,
we again use the symmetric KL divergence to match the mixture components,
and evaluate the MSE on $WW^\top$ to avoid the non-identifiability w.r.t. orthonormal transforms.
%
 To verify the convergence rates w.r.t. $n$, we fix $d$ and let the total dataset $N$ be very large so that $N^{-1}$ is negligible.
 %
Figure~\ref{fig:simple} shows the results when we vary $n$,
where we can see that the MSE of
KL-naive ${\boldsymbol{\hat\theta}}_{\KL}$ is $O(n^{-1})$
while that of KL-control $\hat{\boldsymbol{\theta}}_{\KL-C}$ and KL-weighted $\hat{\boldsymbol{\theta}}_{\KL-W}$ are $O(n^{-2})$;
both are consistent with our results in \eqref{globalmse}, \eqref{globalklc} and \eqref{globalklw}.

\begin{figure}[h]
\begin{centering}
\begin{tabular}{cccc}
\hspace{-3mm}
\includegraphics[width=0.23\textwidth]{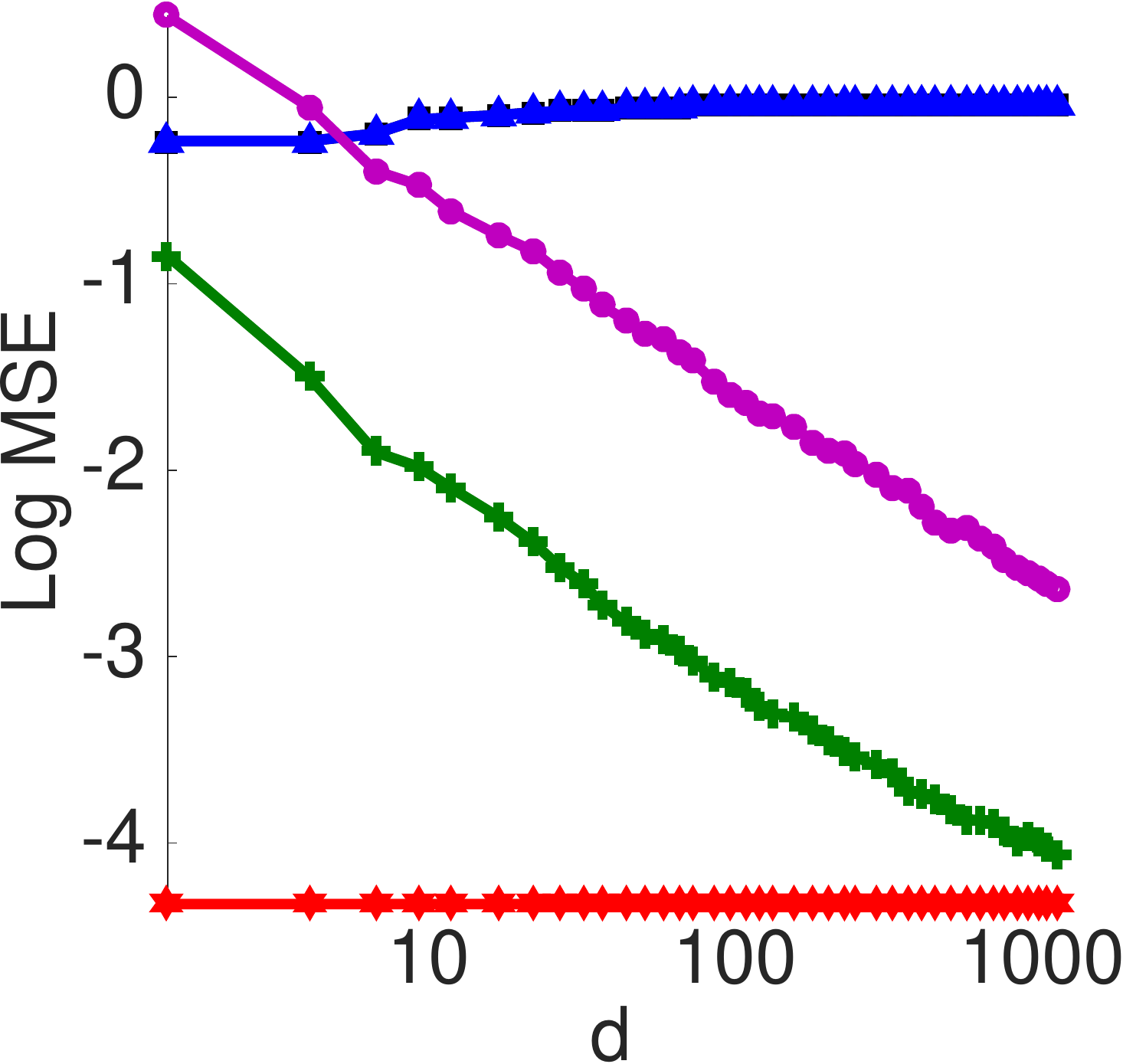} &\hspace{-2mm}
\includegraphics[width=0.23\textwidth]{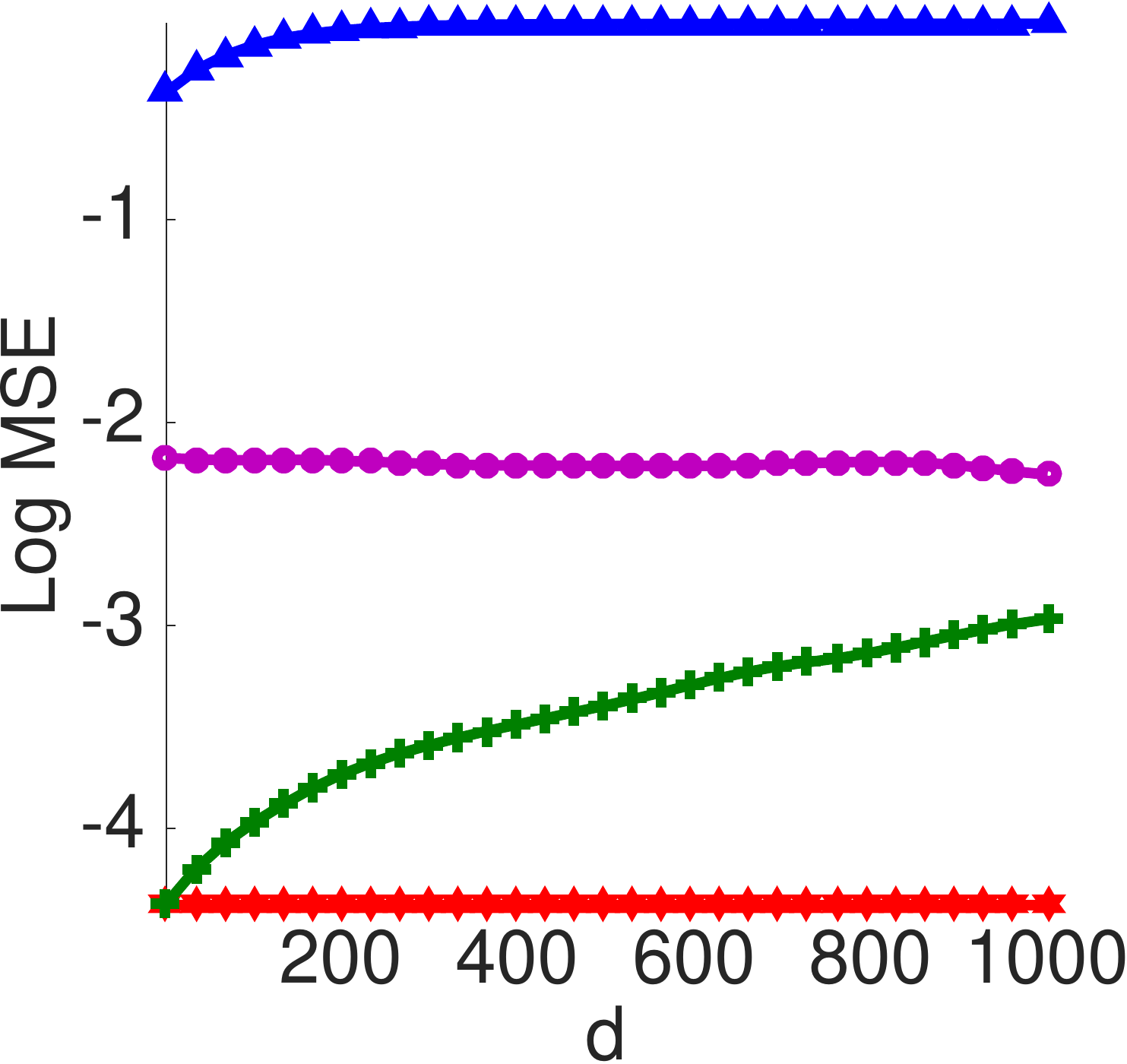} &\hspace{-6mm}
\includegraphics[width=0.23\textwidth]{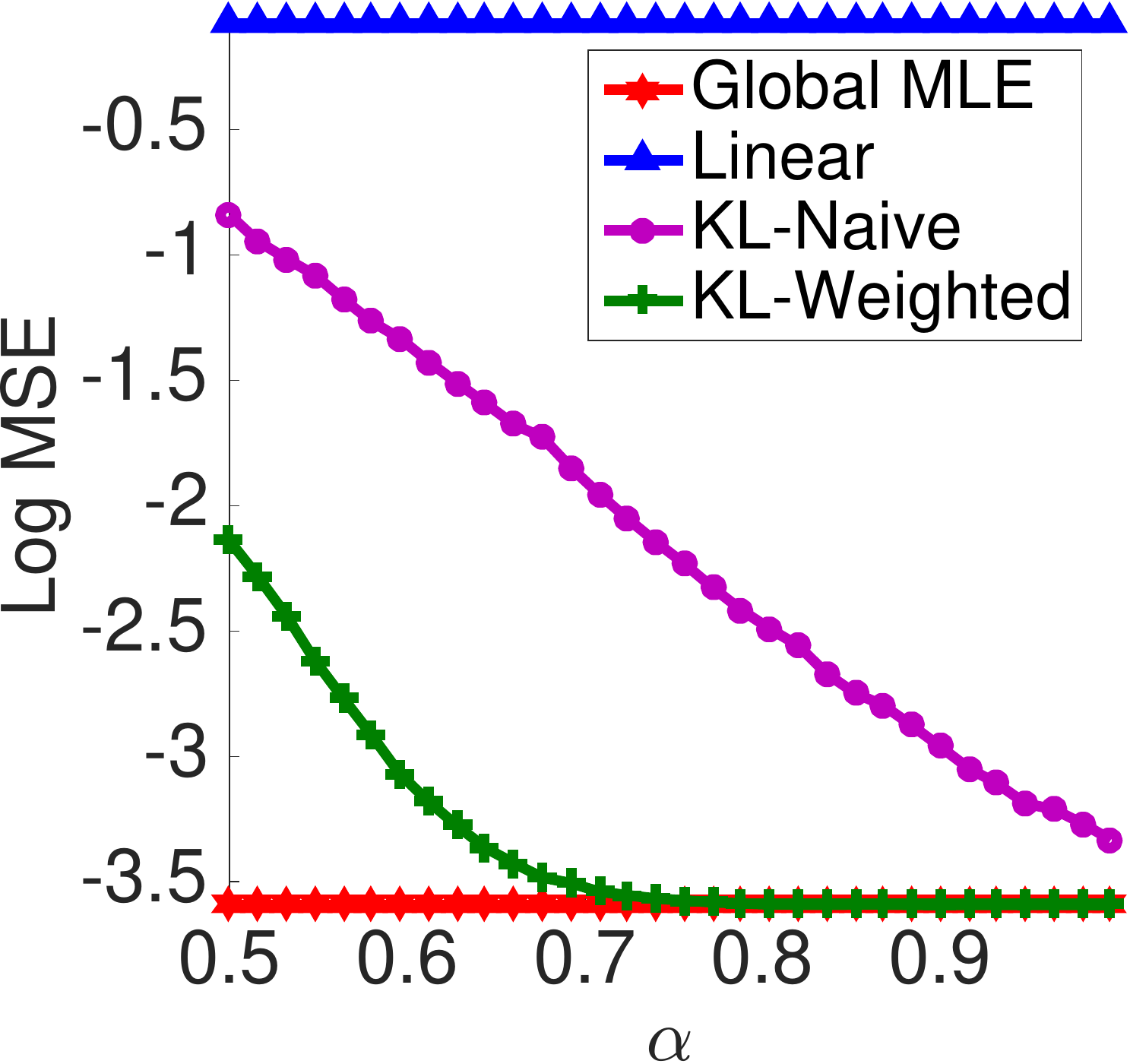} &\hspace{-5mm}
\includegraphics[width=0.23\textwidth]{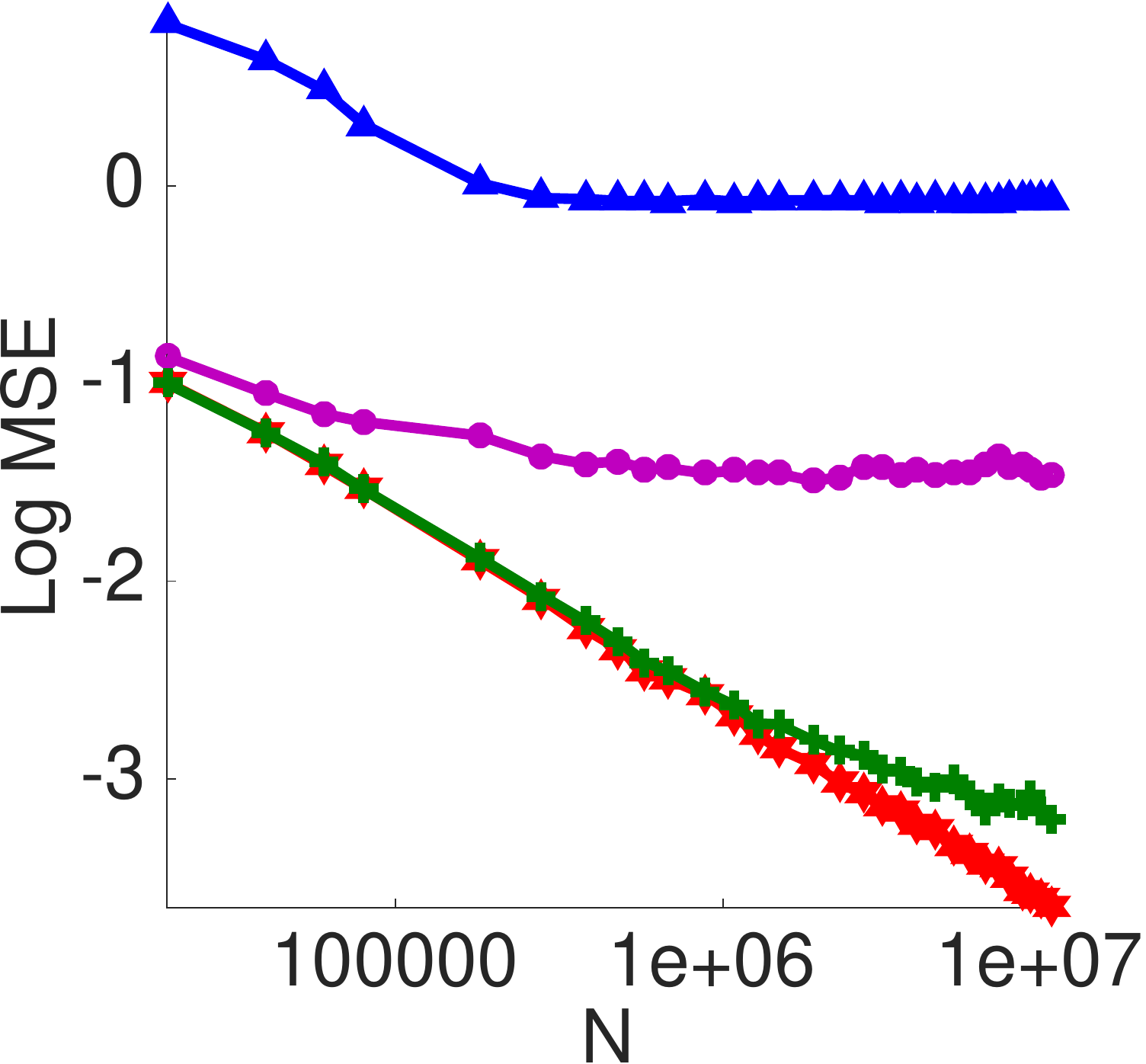} \\
{\small(a) Fix $N$ and $n$} &
{\small(b) Fix $N$ and $n_{\mathrm{tot}}$} &
{\small(c) Fix $N$, $n=(\frac Nd)^{\alpha}$ and $d$} &
 {\small(d) Fix $n$ and $d$}
 \end{tabular}
\caption{Further experiments on PPCA with simulated data.
(a) varying $n$ with fixed $N=5\times 10^7$. (b) varying $d$ with $N=5\times10^7$, $n_{\mathrm{tot}} = n \times d = 3\times 10^5$.
(c) varying $\alpha$ with  $n=(\frac Nd)^{\alpha}$, $N=10^7$ and $d$. (d) varying $N$ with $n=10^3$ and $d = 20$.
The dimension of data $\boldsymbol{x}$ is 5 and the dimension of latent variables $\boldsymbol{t}$ is 4.}
\label{fig:ppcamore}
\end{centering}
\end{figure}

In Figure~\ref{fig:ppcamore}(a),
we increase the number $d$ of local machines,
while using a fix $n$ and a very large $N$,
and find that both $\vv{\hat\theta}_{\mathrm{KL}}$ and $\vv{\hat\theta}_{\mathrm{KL}-W}$ scales as $O(d^{-1})$ as expected.
Note that since the total observation data size $N$ is fixed, the number of data in each local machine is $(N/d)$ and it decreases as we increase $d$.
It is interesting to see that the performance of the KL-based methods actually increases with more partitions;
this is, of course, with a cost of increasing the total bootstrap sample size $d n$ as $d$ increases.
Figure~\ref{fig:ppcamore}(b) considers a different setting,
in which we increase $d$ when fixing the total observation data size $N$, and the total bootstrap sample size $n_{\mathrm{tot}}=n\times d$.
According to \eqref{globalmse} and  \eqref{globalklw}, the MSEs of
$\vv{\hat\theta}_{\mathrm{KL}}$ and $\vv{\hat\theta}_{\mathrm{KL}-W}$ should be about $O(n_{\mathrm{tot}}^{-1})$ and $O(d n_{\mathrm{tot}}^{-2})$ respectively when $N$ is very large, and this is consistent with the results in Figure~\ref{fig:ppcamore}(b).
It is interesting to note that the MSE of $\vv{\hat\theta}_{\mathrm{KL}}$ is independent with $d$ while that of $\vv{\hat\theta}_{\mathrm{KL}-W}$ increases linearly with $d$.
This is not conflict with the fact that $\vv{\hat\theta}_{\mathrm{KL}-W}$ is better than $\vv{\hat\theta}_{\mathrm{KL}}$,  since we always have $d \leq n_{\mathrm{tot}}$.

Figure~\ref{fig:ppcamore}(c) shows the result when we set $n = (N/d)^{\alpha}$ and vary $\alpha$,
where we find that $\vv{\hat\theta}_{\mathrm{KL}-W}$ quickly converges to the global MLE as $\alpha$ increases, while the KL-naive estimator $\vv{\hat\theta}_{\mathrm{KL}}$ converges significantly slower. 
Figure~\ref{fig:ppcamore}(d) demonstrates the case when we increase $N$ while fix $d$ and $n$,
where we see our KL-weighted estimator $\vv{\hat\theta}_{\mathrm{KL}-W}$ matches closely with $N$, except when $N$ is very large in which case the $O((dn^2)^{-1})$ term starts to dominate, while KL-naive is much worse.
We also find the linear averaging estimator performs poorly, and does not scale with $O(N^{-1})$ as the theoretical rate claims;
this is due to unidentifiable orthonormal transform in the PPCA model that we test on.

\subsection{Gaussian Mixture with Unknown Number of Components}
\begin{figure}[h]
\begin{centering}
\begin{tabular}{ccc}
\includegraphics[height=0.24\textwidth]{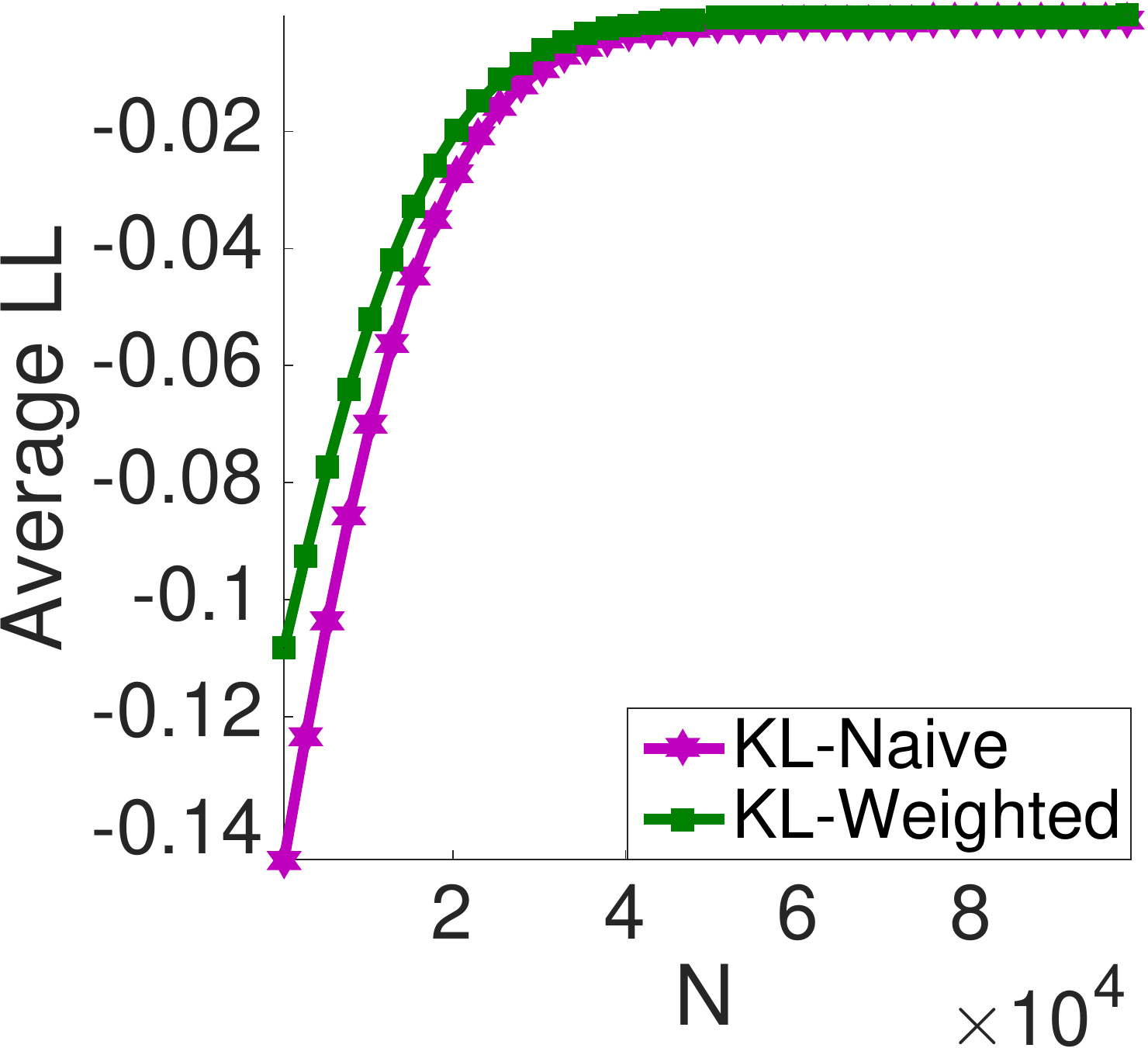} &
\includegraphics[height=0.24\textwidth]{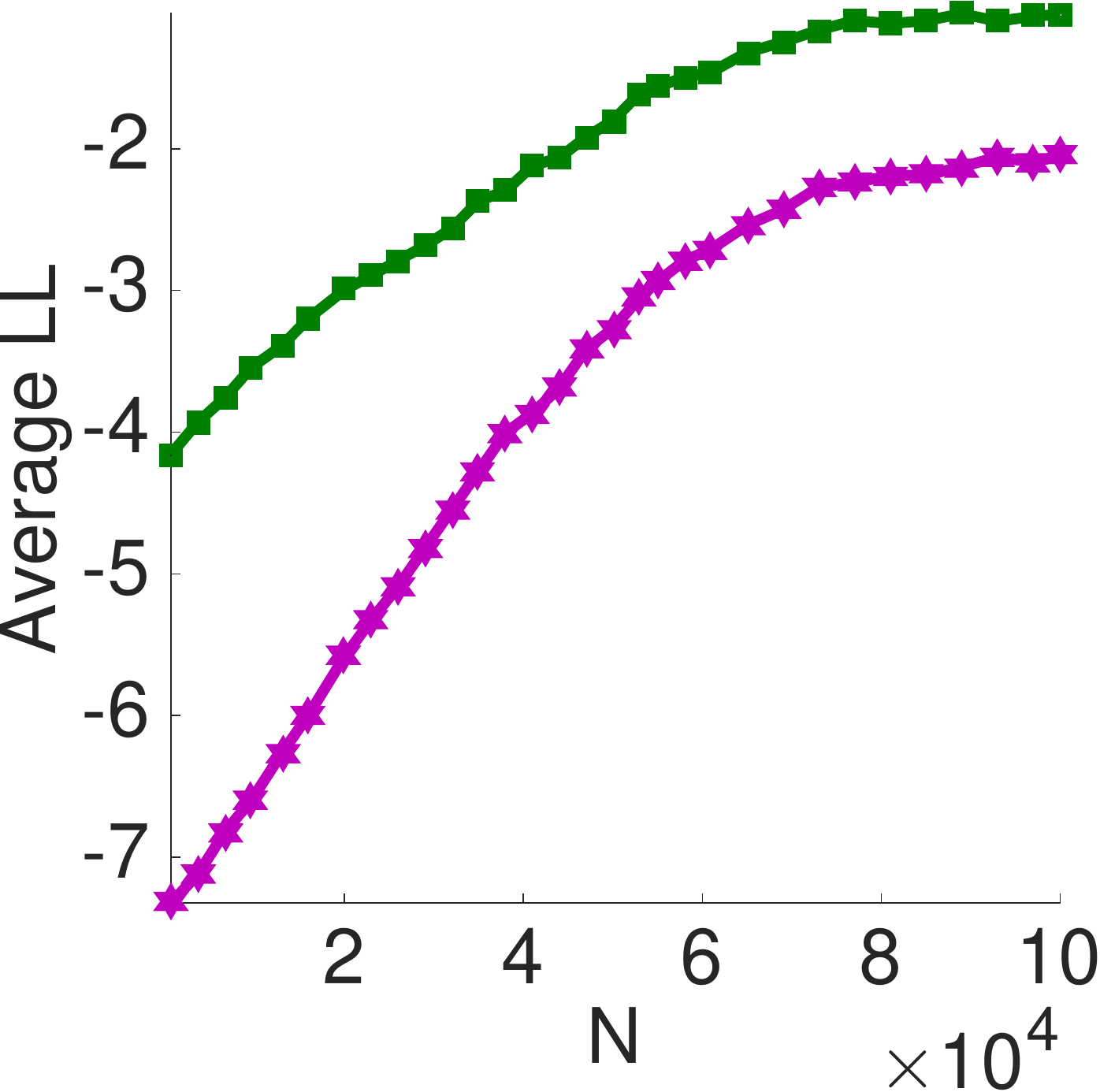} &
\includegraphics[height=0.24\textwidth]{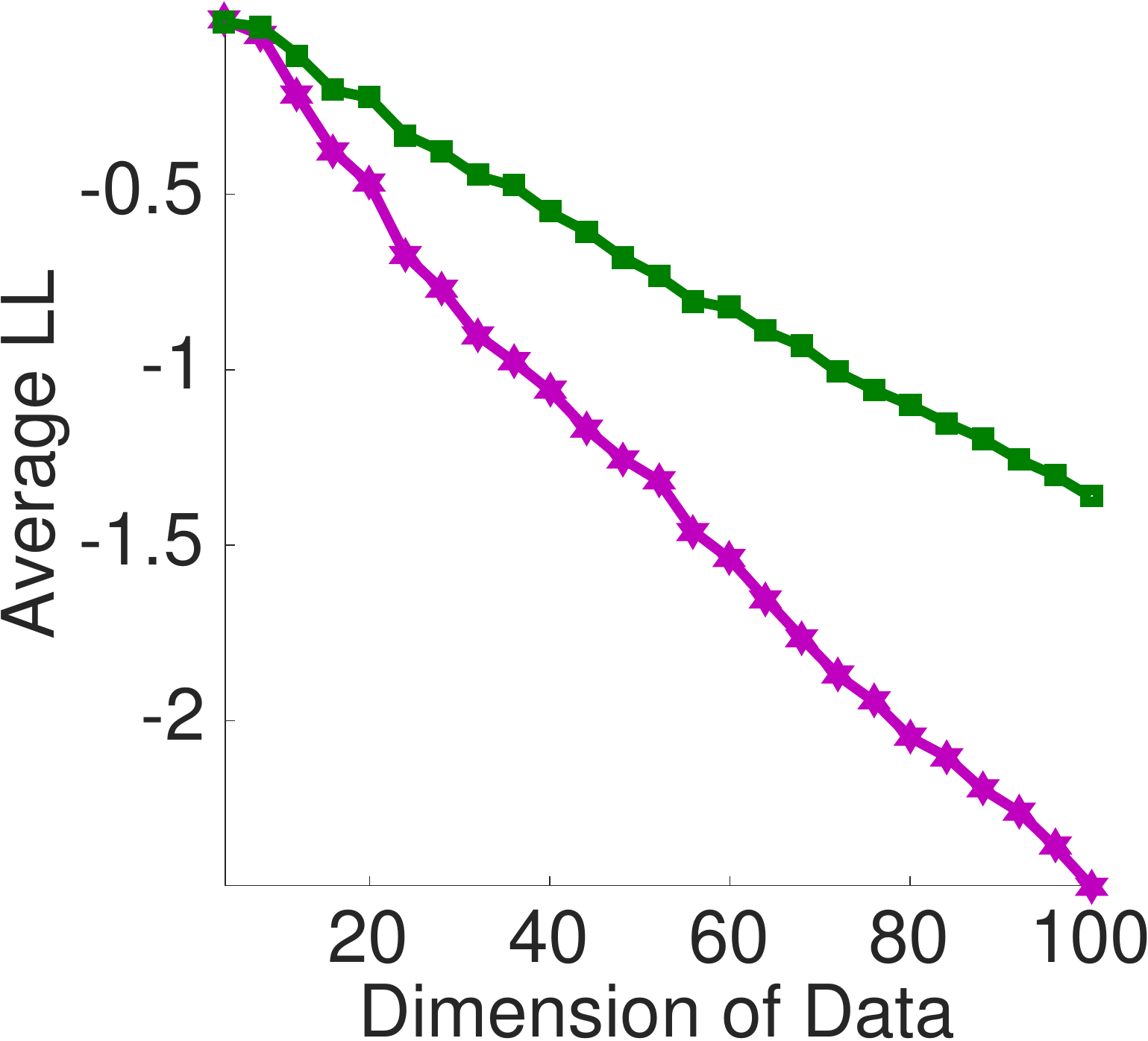} \\
{\small (a)  Dimension of $\vv x$ = 3 } &
{\small (b)  Dimension of $\vv x$ = 80  } &
{\small (c) varying the dimension of $\vv x$}
\end{tabular}
\caption{GMM with the number of mixture components estimated by BIC. We set $n=600$
and the true number of mixtures to be 10 in all the cases.
(a)-(b) vary the total data size $N$ when the dimension of $\vv x$ is 3 and 80, respectively.
(c) varies the dimension of the data with fixed $N=10^5$.
The y-axis is the testing $\log$ likelihood compared with that of global MLE.
}
\label{fig:fig3}
\end{centering}
\end{figure}

We further apply our methods to a more challenging setting for
distributed learning of GMM when the number of mixture components is unknown.
In this case, we first learn each local model with EM and decide its number of components using BIC selection.
Both linear averaging and KL-control $\vv{\hat\theta}_{\KL-C}$ are not applicable in this setting, and and we only test KL-naive $\vv{\hat\theta}_{\mathrm{KL}}$ and KL-weighted $\vv{\hat\theta}_{\mathrm{KL}-W}$.
Since the MSE is also not computable due to the different dimensions, we evaluate
$\vv{\hat\theta}_{\mathrm{KL}}$ and
$\vv{\hat\theta}_{\mathrm{KL}-W}$
using the log-likelihood on a hold-out testing dataset as shown in Figure~\ref{fig:fig3}.
We can see that $\vv{\hat\theta}_{\mathrm{KL}-W}$ generally outperforms
$\vv{\hat\theta}_{\mathrm{KL}}$ as we expect, and the relative improvement increases significantly as the dimension of the observation data $\vv x$ increases.
This suggests that our variance reduction technique works very efficiently in high dimension problems.

\subsection{Results on Real World Datasets}
\begin{figure}[h]
\begin{centering}
\begin{tabular}{ccc}
\includegraphics[height=0.24\textwidth]{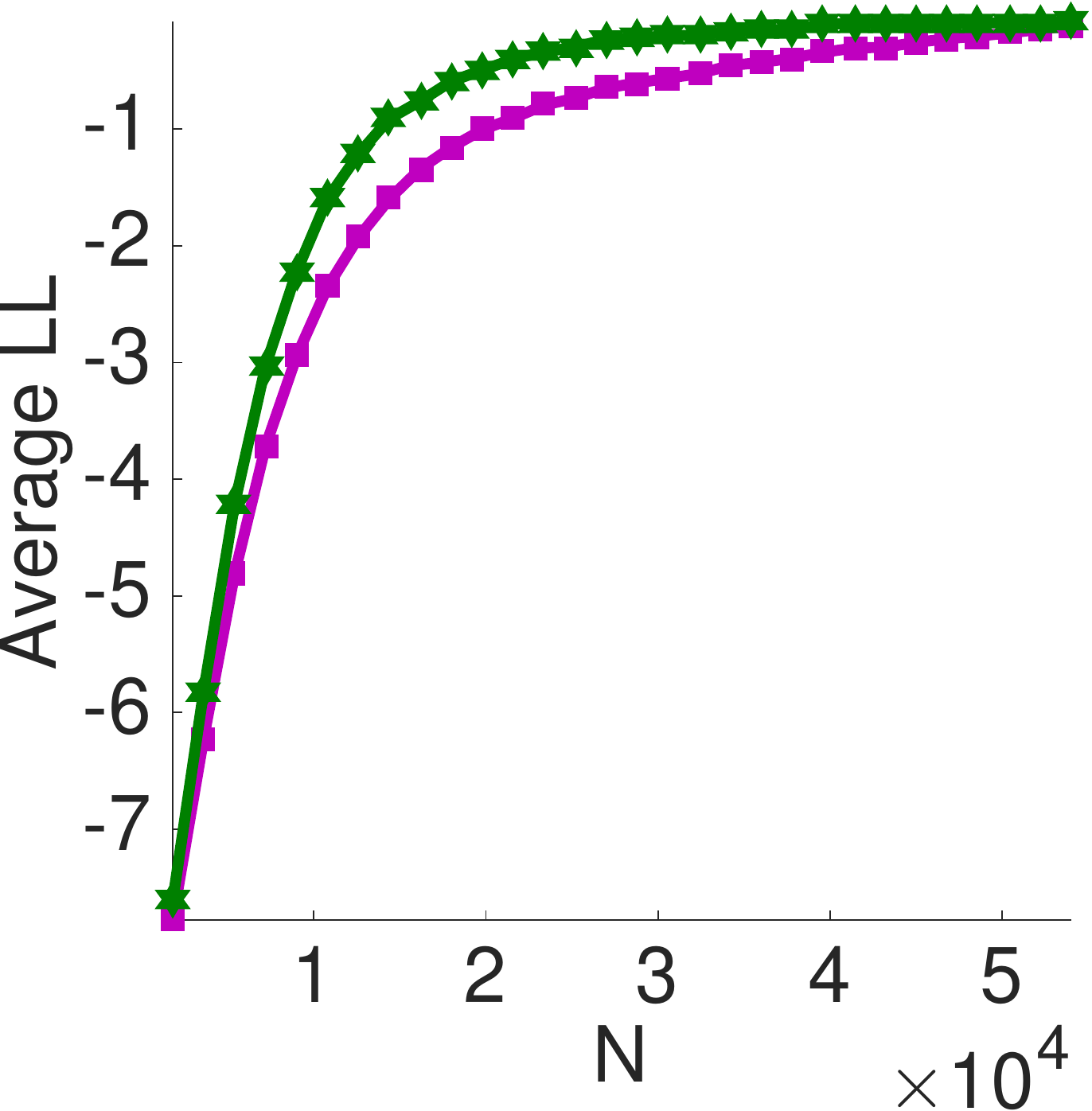} &
\includegraphics[height=0.24\textwidth]{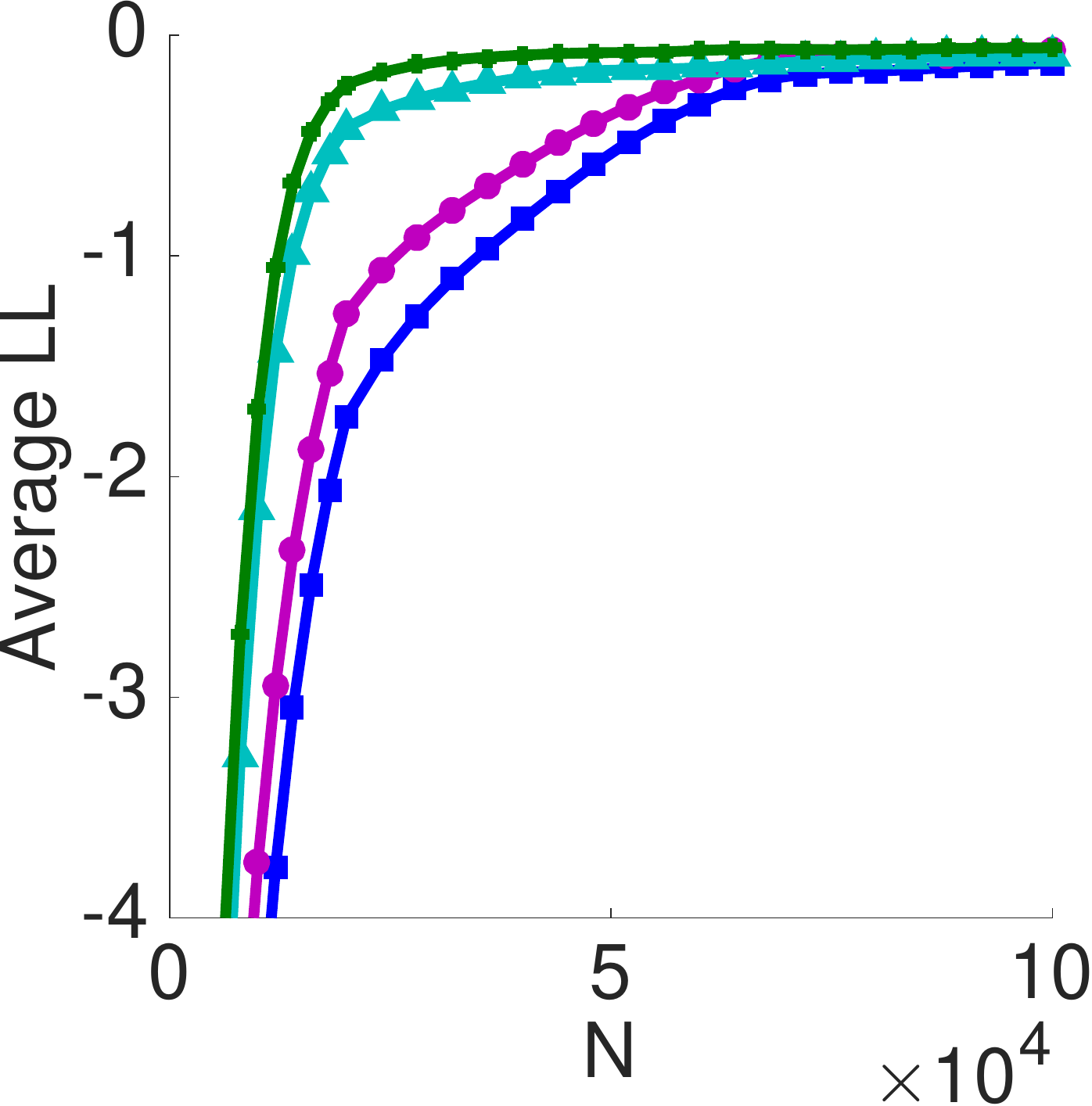} &
\includegraphics[height=0.24\textwidth]{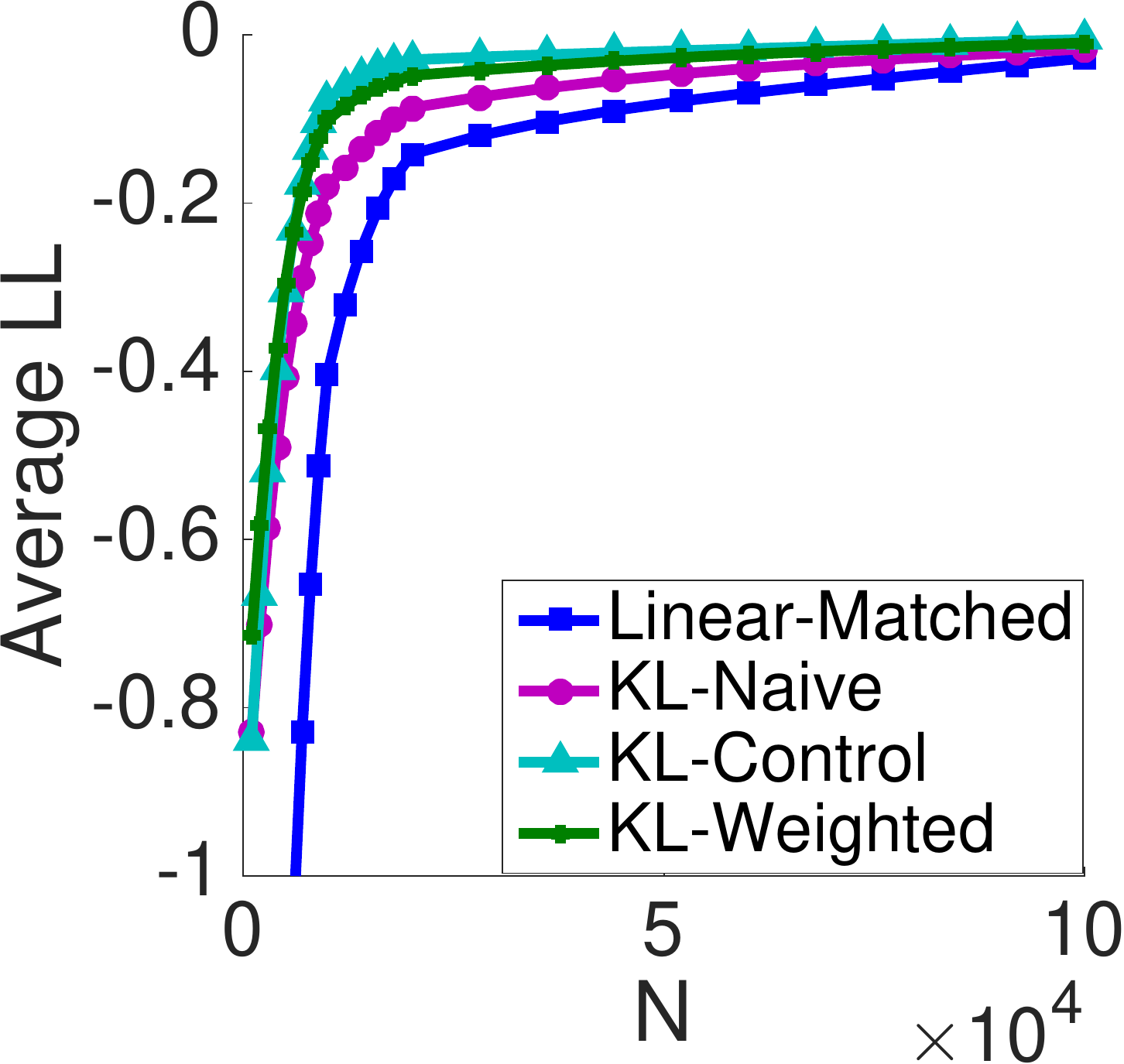} \\
{\small (a) Mixture of PPCA, SensIT Vehicle} &
{\small (b) GMM, Covertype} &
{\small (c) GMM, Epsilon}
\end{tabular}
\caption{Testing $\log$ likelihood (compared with that of global MLE) on real world datasets.
(a) Learning Mixture of PPCA on SensIT Vehicle. (b)-(c) Learning GMM on Covertype and Epsilon.
The number of local machines is 10 in all the cases,
and the number of mixture components are taken to be the number of labels in the datasets. 
The dimension of latent variables in (a) is 90.
For Epsilon, a PCA is first applied and the top 100 principal components are chosen.
Linear-matched and $\KL$-Control are not applicable on Mixture of PPCA and are not shown on (a).
}
\label{fig:fig4}
\end{centering}
\end{figure}

Finally, we apply our methods to several real word datasets, including
the SensIT Vehicle dataset on which mixture of PPCA is tested, and
the Covertype and Epsilon datasets on which GMM is tested.
From Figure~\ref{fig:fig4}, we can see that our KL-Weight  and KL-Control (when it is applicable) again
perform the best. The (matched) linear averaging performs poorly on GMM (Figure~\ref{fig:fig4}(b)-(c)), while is not applicable on mixture of PPCA.


%
%
%
\section{Conclusion and Discussion}
\label{sec:conclusion}
We propose two variance reduction techniques for distributed learning of complex probabilistic models,
including a KL-weighted estimator that is both statistically efficient and widely applicable for even challenging practical scenarios.
Both theoretical and empirical analysis is provided to demonstrate our methods.
Future directions include extending our methods to discriminant learning tasks, as well as the more challenging deep generative networks
on which the exact MLE is not computable tractable, and surrogate likelihood methods with stochastic gradient descent are need.  
We note that the same KL-averaging problem also appears in the ``knowledge distillation" problem in Bayesian deep neural networks \citep{korattikara2015bayesian},
 and it seems that our technique can be applied straightforwardly.

{\bf Acknowledgement} This work is supported in part by NSF CRII 1565796.

\newpage
\bibliographystyle{myunsrtnat}
\bibliography{ref}
\section{Appendix A}

We study the asymptotic property of the KL-naive estimator $\vv {\hat\theta}_{\KL}$, and prove Theorem~\ref{thm1}.

\subsection{Notations and Assumptions}
To simplify the notations for the proofs in the following, we define the following notations.
\begin{equation}
\label{notation}
\begin{aligned}
&s(\boldsymbol{x};\boldsymbol{\theta})=\log p(\boldsymbol{x}\mid\boldsymbol{\theta});\quad \text{\.{s}}(\boldsymbol{x};\boldsymbol{\theta})=\frac{\partial \log p(\boldsymbol{x}\mid\boldsymbol{\theta})}{\partial \boldsymbol{\theta}};\quad \text{\"{s}}(\boldsymbol{x};\boldsymbol{\theta})=\frac{\partial^2 \log p(\boldsymbol{x}\mid\boldsymbol{\theta})}{\partial \boldsymbol{\theta}^2};\\
&I(\boldsymbol{\theta})=\mathbb{E}(\text{\"{s}}(x,\boldsymbol{\theta})); \quad I(\boldsymbol{\hat{\theta}}_k,\boldsymbol{\theta}_{\KL}^*)=\mathbb{E} (\text{\"{s}}(\boldsymbol{x},\boldsymbol{\theta}_{\KL}^*)\mid \boldsymbol{\hat{\theta}}_k).
\end{aligned}
\end{equation}
We start with investigating the theoretical property of $\boldsymbol{\hat{\theta}}_{\KL}$.
\begin{lem}
 Based on Assumption $\ref{assump}$, as $n\to\infty,$ we have $\mathbb{E}(\boldsymbol{\hat{\theta}}_{\KL}-\boldsymbol{\theta}_{\KL}^*)=o((dn)^{-1}).$  Further, in terms of estimating the true parameter, we have
\begin{equation}
\label{globalklc}
\mathbb{E}\|\boldsymbol{\hat{\theta}}_{\KL}-\boldsymbol{\theta}^*\|^2=O(N^{-1}+(dn)^{-1}).
\end{equation}
\end{lem}
{\bf Proof:} Based on Equation (\ref{KLdivmax}) and (\ref{KLdivmaxapprox}), we know
\begin{equation}
\label{KLdiffen}
\sum_{k=1}^d\frac{1}{n}\sum_{j=1}^n \text{\.{s}}(\boldsymbol{\widetilde{x}}_j^k;\boldsymbol{\hat{\theta}}_{\KL})- \sum_{k=1}^d \int p(x|\boldsymbol{\hat{\theta}}_k)\text{\.{s}}(\boldsymbol{x};\boldsymbol{\theta}_{\KL}^*) d\boldsymbol{x}=0.
\end{equation}
By the law of large numbers, we can rewrite Equation (\ref{KLdiffen}) as
\begin{equation}
\label{KLdiffer}
\sum_{k=1}^d \int p(\boldsymbol{x}| \boldsymbol{\hat{\theta}}_k)\text{\.{s}}(\boldsymbol{x}; \boldsymbol{\hat{\theta}}_{\KL})d\boldsymbol{x}-\sum_{k=1}^d \int p(x| \boldsymbol{\hat{\theta}}_k)\text{\.{s}}(\boldsymbol{x};\boldsymbol{\theta}_{\KL}^*)d\boldsymbol{x}=o_p(\frac1n).
\end{equation}
We also observe that $\text{\.{s}}(\boldsymbol{x}; \boldsymbol{\hat{\theta}}_{\KL})-\text{\.{s}}(\boldsymbol{x};\boldsymbol{\theta}_{\KL}^*)= \big[ \int_0^1 \text{\"{s}}(\boldsymbol{x};\boldsymbol{\theta}_{\KL}^*+t(\boldsymbol{\hat{\theta}}_{\KL}-\boldsymbol{\theta}_{\KL}^*))dt \big] ~ (\boldsymbol{\theta}_{\KL}^*-\boldsymbol{\hat{\theta}}_{\KL}).$
Therefore,  Equation (\ref{KLdiffer}) can be written as
\begin{equation}
\label{KLdifferen}
\bigg [\sum_{k=1}^d \int p(x| \boldsymbol{\hat{\theta}}_k)\int_0^1 \text{\"{s}}(\boldsymbol{x};\boldsymbol{\theta}_{\KL}^*+t(\boldsymbol{\hat{\theta}}_{\KL}-\boldsymbol{\theta}_{\KL}^*)) dt d\boldsymbol{x} \bigg] ~ (\boldsymbol{\theta}_{\KL}^*-\boldsymbol{\hat{\theta}}_{\KL}) = o_p(\frac1n).
\end{equation}
Under our Assumption \ref{assump}, the Fish Information matrix $I(\boldsymbol{\theta})$ is positive definite in a neighborhood of $\boldsymbol{\theta}^*,$ then we can find constant $C_1$, $C_2$ such that $C_1\leq\|\int p(x| \boldsymbol{\hat{\theta}}_k)\int_0^1 \text{\"{s}}(\boldsymbol{x};\boldsymbol{\theta}_{\KL}^*+t(\boldsymbol{\hat{\theta}}_{\KL}-\boldsymbol{\theta}_{\KL}^*)) dtd\boldsymbol{x}\|\leq C_2$.  Therefore, we can get $\mathbb{E}(\boldsymbol{\hat{\theta}}_{\KL}-\boldsymbol{\theta}_{\KL}^*)=o((dn)^{-1}).$ $\square$

 The following theorem provides the MSE between $\boldsymbol{\hat{\theta}}_{\KL}$ and $\boldsymbol{\theta}_{\KL}^*$ and that between $\boldsymbol{\hat{\theta}}_{\KL}$ and $\boldsymbol{\theta}^*$.
\begin{thm}
Based on Assumption $\ref{assump}$, as $n\to\infty$, $\mathbb{E}\|\boldsymbol{\hat{\theta}}_{\KL}-\boldsymbol{\theta}_{\KL}^*\|^2=O(\frac{1}{nd}).$ Further, in terms of estimating the true parameter, we have
\begin{equation}
\mathbb{E}\|\boldsymbol{\hat{\theta}}_{\KL}-\boldsymbol{\theta}^*\|^2=O(N^{-1}+(dn)^{-1}).
\end{equation}
\end{thm}
{\bf Proof:} According to the Equation (\ref{KLdivmaxapprox}),
\begin{equation}
\label{unn}
\boldsymbol{\hat{\theta}}_{\KL}=\argmax_{\boldsymbol{\theta}\in\Theta}\sum_{k=1}^d\frac{1}{n}\sum_{j=1}^n s(\boldsymbol{\widetilde{x}}_j^k;\boldsymbol{\theta}).
\end{equation}
Then the first order derivative of Equation (\ref{unn}) with respect to $\boldsymbol{\theta}$ at $\boldsymbol{\theta}=\boldsymbol{\hat{\theta}}_{\KL}$ is zero,
\begin{equation}
\label{taylorapp}
\sum_{k=1}^d\frac{1}{n}\sum_{j=1}^n \text{\.{s}}(\boldsymbol{\widetilde{x}}_j^k;\boldsymbol{\hat{\theta}}_{\KL})=0.
\end{equation}
By Taylor expansion of Equation (\ref{taylorapp}), we get
$$\sum_{k=1}^d\frac{1}{n}\sum_{j=1}^n (\text{\.{s}}(\boldsymbol{\widetilde{x}}_j^k;\boldsymbol{\theta}_{\KL}^*)+\text{\"{s}}(\boldsymbol{\widetilde{x}}_j^k;\boldsymbol{\hat{\theta}}_{\KL})(\boldsymbol{\hat{\theta}}_{\KL}-\boldsymbol{\theta}_{\KL}^*))+o_p(\boldsymbol{\hat{\theta}}_{\KL}-\boldsymbol{\theta}_{\KL}^*)=0.$$
By the law of large numbers, $\frac{1}{n}\sum_{j=1}^n\text{\"{s}}(\boldsymbol{\widetilde{x}}_j^k;\boldsymbol{\hat{\theta}}_{\KL}^*)=I(\boldsymbol{\hat{\theta}}_k,\boldsymbol{\theta}_{\KL}^*)+o_p(\frac{1}{n}).$ 
Under our Assumption \ref{assump},  $I(\boldsymbol{\theta})$ is positive definite in a neighborhood of $\boldsymbol{\theta}^*.$ Since $\hat{\boldsymbol{\theta}}_k$ are in the neighborhood of $\boldsymbol{\theta}^*$, $I(\boldsymbol{\hat{\theta}}_k, \boldsymbol{\theta}_{\KL}^*)$ is positive definite, for $k=1\in [d].$ Then we have
\begin{equation}
\label{varianceKL}
\boldsymbol{\hat{\theta}}_{\KL}-\boldsymbol{\theta}_{\KL}^*=(\sum_{k=1}^dI(\boldsymbol{\hat{\theta}}_k,\boldsymbol{\theta}_{\KL}^*))^{-1}\sum_{k=1}^d\frac{1}{n}\sum_{j=1}^n \text{\.{s}}(\boldsymbol{\widetilde{x}}_j^k;\boldsymbol{\theta}_{\KL}^*)+o_p(\frac{1}{n})=0.
\end{equation}
By the central limit theorem, $\frac{1}{\sqrt{n}}\sum_{j=1}^n \text{\.{s}}(\boldsymbol{\widetilde{x}}_j^k;\boldsymbol{\theta}_{\KL}^*)$ converges to a normal distribution. By some simple calculation, we have
\begin{equation}
\label{cova}
\mathrm{Cov}(\boldsymbol{\hat{\theta}}_{\KL}-\boldsymbol{\theta}_{\KL}^*,\boldsymbol{\hat{\theta}}_{\KL}-\boldsymbol{\theta}_{\KL}^*)=\frac{1}{n}(\sum_{k=1}^dI(\boldsymbol{\hat{\theta}}_k,\boldsymbol{\theta}_{\KL}^*))^{-1}\sum_{k=1}^d \mathrm{Var}(\text{\.{s}}(\boldsymbol{x};\boldsymbol{\theta}_{\KL}^*)\mid\boldsymbol{\hat{\theta}}_k)(\sum_{k=1}^dI(\boldsymbol{\hat{\theta}}_k,\boldsymbol{\theta}_{\KL}^*))^{-1}.
\end{equation}
According to our Assumption \ref{assump}, we already know $I(\boldsymbol{\hat{\theta}}_k,\boldsymbol{\theta}_{\KL}^*)$ is positive definite, $C_1\le \|I(\boldsymbol{\hat{\theta}}_k,\boldsymbol{\theta}_{\KL}^*)\|\le C_2$. We have $(\sum_{k=1}^dI(\boldsymbol{\hat{\theta}}_k,\boldsymbol{\theta}_{\KL}^*))^{-1}=O(\frac{1}{d})$ and $\sum_{k=1}^d \mathrm{Var}(\text{\.{s}}(\boldsymbol{x};\boldsymbol{\theta}_{\KL}^*)\mid\boldsymbol{\hat{\theta}}_k)=O(d).$
Therefore,  $\mathbb{E}\|\boldsymbol{\hat{\theta}}_{\KL}-\boldsymbol{\theta}_{\KL}^*\|^2=\mathrm{trace}(\mathrm{Cov}(\boldsymbol{\hat{\theta}}_{\KL}-\boldsymbol{\theta}_{\KL}^*,\boldsymbol{\hat{\theta}}_{\KL}-\boldsymbol{\theta}_{\KL}^*))=O(\frac{1}{nd}).$ Because the MSE between the exact $\KL$ estimator $\boldsymbol{\theta}_{\KL}^*$ and the true parameter $\boldsymbol{\theta}^*$ is
$O(N^{-1})$
as shown in \citet{liu2014distributed}, the MSE between $\boldsymbol{\hat{\theta}}_{\KL}$ and the true parameter $\boldsymbol{\theta}^*$ is
\begin{equation*}
\mathbb{E}\|\boldsymbol{\hat{\theta}}_{\KL}-\boldsymbol{\theta}^*\|^2\approx\mathbb{E}\|\boldsymbol{\hat{\theta}}_{\KL}-\boldsymbol{\theta}^*_{\KL}\|^2+\mathbb{E}\|\boldsymbol{\theta}_{\KL}^*-\boldsymbol{\theta}^*\|^2=O(N^{-1}+(dn)^{-1}).
\end{equation*}
We complete the proof of this theorem. $\square$
\section{Appendix B}
In this section, we analyze the MSE of our proposed estimator $\boldsymbol{\hat{\theta}}_{\KL-C}$ and prove Theorem~\ref{Control}.
\begin{thm}
Under Assumptions \ref{assump}, we have
$$\text{as }~ n\to \infty,\quad n\mathbb{E}\|\boldsymbol{\hat{\theta}}_{\KL-C}-\boldsymbol{\theta}_{\KL}^*\|^2 < n\mathbb{E}\|\boldsymbol{\hat{\theta}}_{\KL}-\boldsymbol{\theta}_{\KL}^*\|^2.$$
\end{thm}
Since $\widetilde{\boldsymbol{\theta}}_k$ is the MLE of data $\{\boldsymbol{\widetilde{x}}_j^k\}_{j=1}^n$, then we have
\begin{equation}
\label{MLE}
(\boldsymbol{\widetilde{\theta}}_k-\boldsymbol{\hat{\theta}}_k)= -I(\boldsymbol{\hat{\theta}}_k)^{-1}\frac{1}{n}\sum_{j=1}^n\dot{s}(\boldsymbol{\widetilde{x}}_j^k;\boldsymbol{\hat{\theta}}_k)+o_p(\frac{1}{n}).
\end{equation}
Then $\mathbb{E}(\boldsymbol{\widetilde{\theta}}_k-\boldsymbol{\hat{\theta}}_k)=o(\frac{1}{n}).$ According to Theorem (\ref{thm1}), when $\boldsymbol{\mathfrak{B}}_k$ is a constant matrix, for $k\in[d],$
$$\mathbb{E}(\boldsymbol{\hat{\theta}}_{\KL-C}-\boldsymbol{\theta}_{\KL}^*)=\mathbb{E}(\boldsymbol{\hat{\theta}}_{\KL}-\boldsymbol{\theta}_{\KL}^*)+\sum_{k=1}^d \boldsymbol{\mathfrak{B}}_k\mathbb{E}(\boldsymbol{\widetilde{\theta}}_k-\boldsymbol{\hat{\theta}}_k)=o(\frac{1}{n}).$$

Notice that $\frac{1}{n}\sum_{j=1}^n\text{\.{s}}(\boldsymbol{\widetilde{x}}_j^r;\boldsymbol{\hat{\theta}}_{r})$ and $\frac{1}{n}\sum_{j=1}^n\text{\.{s}}(\boldsymbol{\widetilde{x}}_j^t;\boldsymbol{\hat{\theta}}_{t})$ are independent when $r\neq t.$
According to Equation (\ref{varianceKL}), we know $\sum_{k=1}^d\frac1n\sum_{j=1}^n \text{\.{s}}(\boldsymbol{\widetilde{x}}_j^k;\boldsymbol{\theta}_{\KL}^*)$ and $\frac{1}{n}\sum_{j=1}^n\dot{s}(\boldsymbol{\widetilde{x}}_j^k;\boldsymbol{\hat{\theta}}_k)$ are correlated to each other for $k\in [d],$
\begin{equation*}
\begin{split}
&\mathrm{Cov}((\boldsymbol{\hat{\theta}}_{\KL-C}-\boldsymbol{\theta}_{\KL}^*),(\boldsymbol{\hat{\theta}}_{\KL-C}-\boldsymbol{\theta}_{\KL}^*))=\mathrm{Cov}(\boldsymbol{\hat{\theta}}_{\KL}-\boldsymbol{\theta}_{\KL}^*,\boldsymbol{\hat{\theta}}_{\KL}-\boldsymbol{\theta}_{\KL}^*)\\
&+2\sum_{k=1}^d\boldsymbol{\mathfrak{B}}_k
\mathrm{Cov}(\boldsymbol{\hat{\theta}}_{\KL}-\boldsymbol{\theta}_{\KL},\boldsymbol{\widetilde{\theta}}_k-\hat{\boldsymbol{\theta}}_{k})^T+
\sum_{k=1}^d\boldsymbol{\mathfrak{B}}_k\mathrm{Cov}((\boldsymbol{\widetilde{\theta}}_k-\boldsymbol{\hat{\theta}}_{k}),(\boldsymbol{\widetilde{\theta}}_k-\boldsymbol{\hat{\theta}}_{k}))\boldsymbol{\mathfrak{B}}_k^T.
\end{split}
\end{equation*}
When $\mathfrak{\boldsymbol{B}}_k=-(\mathrm{Cov}(\boldsymbol{\widetilde{\theta}}_k-\boldsymbol{\hat{\theta}}_{k},\boldsymbol{\widetilde{\theta}}_k-\boldsymbol{\hat{\theta}}_{k}))^{-1}\mathrm{Cov}(\boldsymbol{\hat{\theta}}_{\KL}-
\boldsymbol{\theta}_{\KL}^*,\boldsymbol{\widetilde{\theta}}_k-\hat{\boldsymbol{\theta}}_{k}),$
we have
\begin{equation}
\label{controleff}
\begin{split}
&\mathrm{Cov}(\boldsymbol{\hat{\theta}}_{\KL-C}-\boldsymbol{\theta}_{KL}^*,\boldsymbol{\hat{\theta}}_{\KL-C}-\boldsymbol{\theta}_{\KL}^*)=\mathrm{Cov}(\boldsymbol{\hat{\theta}}_{\KL}-\boldsymbol{\theta}_{\KL}^*,\boldsymbol{\hat{\theta}}_{\KL}-\boldsymbol{\theta}_{\KL}^*)-\\
&\sum_{k=1}^d\mathrm{Cov}(\boldsymbol{\widetilde{\theta}}_k-\boldsymbol{\hat{\theta}}_{k},\boldsymbol{\widetilde{\theta}}_k-\boldsymbol{\hat{\theta}}_{k})^{-1}\mathrm{Cov}(\boldsymbol{\hat{\theta}}_{KL}-
\boldsymbol{\theta}_{\KL}^*,\boldsymbol{\widetilde{\theta}}_k-\hat{\boldsymbol{\theta}}_{k})\mathrm{Cov}(\boldsymbol{\hat{\theta}}_{\KL}-\boldsymbol{\theta}_{\KL}^*,\boldsymbol{\widetilde{\theta}}_k-
\boldsymbol{\hat{\theta}}_{k})^T.
\end{split}
\end{equation}
We know $\mathbb{E}\|\boldsymbol{\hat{\theta}}_{\KL-C}-\boldsymbol{\theta}_{\KL}^*\|^2=\mathrm{trace}(\mathrm{Cov}(\boldsymbol{\hat{\theta}}_{\KL-C}-\boldsymbol{\theta}_{\KL}^*,\boldsymbol{\hat{\theta}}_{\KL-C}-\boldsymbol{\theta}_{\KL}^*))$, $ \mathbb{E}\|\boldsymbol{\hat{\theta}}_{\KL}-\boldsymbol{\theta}_{\KL}^*\|^2=\mathrm{trace}(\mathrm{Cov}(\boldsymbol{\hat{\theta}}_{\KL}-\boldsymbol{\theta}_{\KL}^*,\boldsymbol{\hat{\theta}}_{\KL}-\boldsymbol{\theta}_{\KL}^*)).$ The second term of Equation (\ref{controleff}) is a positive definite matrix, therefore we have
$n\mathbb{E}\|\boldsymbol{\hat{\theta}}_{\KL-C}-\boldsymbol{\theta}_{\KL}^*\|^2< n\mathbb{E}\|\boldsymbol{\hat{\theta}}_{\KL}-\boldsymbol{\theta}_{\KL}^*\|^2$ as $n\to\infty.$
We complete the proof of this theorem. $\square$
\begin{thm}
Under Assumption \ref{assump}, when $N> n\times d$, we have $E\|\boldsymbol{\hat{\theta}}_{\KL-C}- \boldsymbol{\theta}_{\KL}^*\|^2=O(\frac{1}{dn^2})$ as $n\to\infty.$ Further, in terms of estimating the true parameter, we have
\begin{equation*}
\label{globalklc}
\mathbb{E}\|\boldsymbol{\hat{\theta}}_{\KL-C}-\boldsymbol{\theta}^*\|^2=O(N^{-1}+(dn^{2})^{-1}).
\end{equation*}
\end{thm}
From Equation (\ref{KLdivmaxapprox}), we know
\begin{equation}
\label{MLEequation}
\sum_{k=1}^d\frac1n\sum_{j=1}^n\frac{\partial \log p(\boldsymbol{\widetilde{x}}_j^k| \boldsymbol{\hat{\theta}}_{\KL})}{\partial \boldsymbol{\theta}}=0.
\end{equation}
By Taylor expansion, Equation (\ref{MLEequation}) can be rewritten as
\begin{equation}
\label{MLEapproximate}
\sum_{k=1}^d[\frac1n\sum_{j=1}^n\text{\.{s}}(\boldsymbol{\widetilde{x}}_j^k;\boldsymbol{\hat{\theta}}_{k})+ \text{\"{s}}(\boldsymbol{\widetilde{x}}_j^k;\hat{\boldsymbol{\theta}}_{k})(\boldsymbol{\hat{\theta}}_{\KL}-\boldsymbol{\hat{\theta}}_{k}))+O_p(\|\boldsymbol{\hat{\theta}}_{\KL}-\boldsymbol{\hat{\theta}}_{k}\|^2)]=0.
\end{equation}
$\|\boldsymbol{\hat{\theta}}_{\KL}-\boldsymbol{\hat{\theta}}_{k}\|^2\leq \|\boldsymbol{\hat{\theta}}_{\KL}-\boldsymbol{\theta}_{\KL}^*\|^2+\|\boldsymbol{\theta}_{\KL}^*-\boldsymbol{\hat{\theta}}_{k}\|^2$. As we know from \citet{liu2014distributed}, we have
\begin{equation}
\label{liuproof}
\|\boldsymbol{\theta}_{\KL}^*-\boldsymbol{\hat{\theta}}_{k}\|^2\le\|\boldsymbol{\theta}_{\KL}^*- \boldsymbol{\theta}^*\|^2+\|\boldsymbol{\theta}^*- \boldsymbol{\hat{\theta}}_k\|^2=O_p(\frac{d}{N}),
\end{equation}
When $N> n\times d$, we have $\|\boldsymbol{\hat{\theta}}_{\KL}-\boldsymbol{\hat{\theta}}_{k}\|^2=O_p(\frac{1}{nd})$. And it is also easy to derive
\begin{equation}
\label{asym}
\boldsymbol{\hat{\theta}}_{\KL}-\boldsymbol{\hat{\theta}}_{k}=\boldsymbol{\hat{\theta}}_{\KL}-\boldsymbol{\theta}_{\KL}^*+\boldsymbol{\theta}_{\KL}^*-\boldsymbol{\theta}^*+\boldsymbol{\theta}^*-\boldsymbol{\hat{\theta}}_{k}=
o_p(\frac{1}{N})+o_p(\frac{1}{N})+o_p(\frac{d}{N})=o_p(\frac{1}{nd}+\frac{d}{N}),
\end{equation}
where $\boldsymbol{\theta}_{KL}^*-\boldsymbol{\theta}^*=o_p(\frac1N)$ has been proved in Liu and Ihler's paper(2014).
According to the law of large numbers, $\frac1n\sum_{j=1}^n\text{\"{s}}(\boldsymbol{\widetilde{x}}_j^k;\boldsymbol{\hat{\theta}}_{k})= I(\boldsymbol{\hat{\theta}}_k)+o_p(\frac{1}{n})$, then we have
\begin{equation}
\label{MLEcalculate}
(\boldsymbol{\hat{\theta}}_{\KL}-\boldsymbol{\theta}_{\KL}^*)=-(\sum_{k=1}^dI(\boldsymbol{\hat{\theta}}_k))^{-1}\sum_{k=1}^d\frac{1}{n}\sum_{j=1}^n\text{\.{s}}(\boldsymbol{\widetilde{x}}_j^k;\boldsymbol{\hat{\theta}}_{k})+
O_p(\frac{1}{nd}).
\end{equation}
Notie that $\frac{1}{n}\sum_{j=1}^n\text{\.{s}}(\boldsymbol{\widetilde{x}}_j^r;\boldsymbol{\hat{\theta}}_{r})$ and $\frac{1}{n}\sum_{j=1}^n\text{\.{s}}(\boldsymbol{\widetilde{x}}_j^t;\boldsymbol{\hat{\theta}}_{t})$ are independent when $r\neq t.$
Therefore from (\ref{MLE}) and (\ref{MLEcalculate}), the covariance matrix of $n(\boldsymbol{\hat{\theta}}_{\KL}-\boldsymbol{\theta}_{\KL}^*)$ and $n(\widetilde{\boldsymbol{\theta}}_k-\boldsymbol{\hat{\theta}}_k)$ is
$$
\mathrm{Cov}(n(\boldsymbol{\hat{\theta}}_{\KL}-\boldsymbol{\theta}_{\KL}^*),n(\boldsymbol{\widetilde{\theta}}_k-\boldsymbol{\hat{\theta}}_{k}))=n(\sum_{k=1}^dI(\boldsymbol{\hat{\theta}}_k))^{-1}+(\sum_{k=1}^dI(\boldsymbol{\hat{\theta}}_k))^{-1}O(1),
$$
for $k\in [d].$ According to Assumption \ref{assump}, we know $\sum_{k=1}^dI(\boldsymbol{\hat{\theta}}_k)=O(d)$. Then we will have 
\begin{equation}
\label{covariance}
\mathrm{Cov}(n(\boldsymbol{\hat{\theta}}_{\KL}-\boldsymbol{\theta}_{\KL}^*),n(\boldsymbol{\widetilde{\theta}}_k-\boldsymbol{\hat{\theta}}_{k}))=n(\sum_{k=1}^dI(\boldsymbol{\hat{\theta}}_k))^{-1}+O(\frac1d), ~~\text{for}~~k\in[d].
\end{equation}

According to Theorem \ref{thm1} and Equation (\ref{cova}), by the law of large numbers, it is easy to derive
$$\mathrm{Cov}(n(\boldsymbol{\hat{\theta}}_{\KL}-\boldsymbol{\theta}_{KL}^*),n(\boldsymbol{\hat{\theta}}_{\KL}-\boldsymbol{\theta}_{\KL}^*))=n(\sum_{k=1}^dI(\boldsymbol{\hat{\theta}}_k))^{-1}+o(1).$$
\begin{equation}
\label{controlcov}
\begin{split}
&\mathrm{Cov}(n(\boldsymbol{\hat{\theta}}_{\KL-C}-\boldsymbol{\theta}_{\KL}^*),n(\boldsymbol{\hat{\theta}}_{\KL-C}-\boldsymbol{\theta}_{\KL}^*))=\mathrm{Cov}(n(\boldsymbol{\hat{\theta}}_{\KL}-\boldsymbol{\theta}_{\KL}^*),n(\boldsymbol{\hat{\theta}}_{\KL}-\boldsymbol{\theta}_{\KL}^*)\\
&+2\sum_{k=1}^d\boldsymbol{\mathfrak{B}}_k
\mathrm{Cov}(n(\boldsymbol{\hat{\theta}}_{\KL}-\boldsymbol{\theta}_{\KL}^*),n(\boldsymbol{\widetilde{\theta}}_k-\boldsymbol{\hat{\theta}}_{k}))^\top+
\sum_{k=1}^d\boldsymbol{\mathfrak{B}}_k\mathrm{Cov}(n(\boldsymbol{\widetilde{\theta}}_k-\boldsymbol{\hat{\theta}}_{k}),n(\boldsymbol{\widetilde{\theta}}_k-\boldsymbol{\hat{\theta}}_{k}))\boldsymbol{\mathfrak{B}}_k^T,
\end{split}
\end{equation}
where $\boldsymbol{\mathfrak{B}}_k$ is defined in (\ref{scorecoeff}),
\begin{equation*}
\boldsymbol{\mathfrak{B}}_k=-(\sum_{k=1}^dI(\boldsymbol{\hat{\theta}}_k))^{-1}I(\boldsymbol{\hat{\theta}}_k), \quad k\in [d].
\end{equation*}

According to Equation (\ref{MLE}), we know $\mathrm{Cov}(n(\boldsymbol{\widetilde{\theta}}_k-\boldsymbol{\hat{\theta}}_{k}),n(\boldsymbol{\widetilde{\theta}}_k-\boldsymbol{\hat{\theta}}_{k}))=n(I(\vv {\hat\theta}_k))^{-1}+o(1).$ By some simple calculation, we know that $n^2\mathrm{Cov}(\boldsymbol{\hat{\theta}}_{\KL-C}-\boldsymbol{\theta}_{\KL}^*,\boldsymbol{\hat{\theta}}_{\KL-C}-\boldsymbol{\theta}_{\KL}^*)=O(\frac{1}{d}).$  Therefore, under the Assumption \ref{assump}, when $N> n\times d,$  we get the following result,
$$
\mathbb{E}\|\boldsymbol{\hat{\theta}}_{\KL-C}- \boldsymbol{\theta}_{\KL}^*\|^2=\mathrm{trace}(\mathrm{Cov}(\boldsymbol{\hat{\theta}}_{\KL-C}-\boldsymbol{\theta}_{\KL}^*,\boldsymbol{\hat{\theta}}_{\KL-C}-\boldsymbol{\theta}_{\KL}^*))=O(\frac{1}{dn^2}).$$
We know $\mathbb{E}\|\boldsymbol{\theta}_{\KL}^*-\vv \theta^*\|^2=O(N^{-1})$ from \citet{liu2014distributed}. Then we have
$$\mathbb{E}\|\boldsymbol{\hat{\theta}}_{\KL-C}-\boldsymbol{\theta}^*\|^2\approx\mathbb{E}\|\boldsymbol{\hat{\theta}}_{\KL-C}-\boldsymbol{\theta}^*_{\KL}\|^2+\mathbb{E}\|\boldsymbol{\theta}_{\KL}^*-\boldsymbol{\theta}^*\|^2=O(N^{-1}+(dn^2)^{-1}).$$
The proof of this theorem is complete. $\square$
\section{Appendix C}
In this section, we analyze the asymptotic property of $\boldsymbol{\hat{\theta}}_{\KL-W}$ and prove Theorem~\ref{thm3}. We show the MSE between $\boldsymbol{\hat{\theta}}_{\KL-W}$ and $\boldsymbol{\theta}_{\KL}^*$ is much smaller than the MSE between the $\KL$-naive estimator $\boldsymbol{\hat{\theta}}_{\KL}$ and $\boldsymbol{\theta}_{\KL}^*.$
\begin{lem}
Under Assumption \ref{assump}, as $n\to \infty$, $\widetilde{\eta}(\boldsymbol{\theta})$ is a more accurate estimator of $\eta(\boldsymbol{\theta})$ than $\hat{\eta}(\boldsymbol{\theta})$, i.e.,
\begin{equation}
n\mathrm{Var}(\widetilde{\eta}(\boldsymbol{\theta}))\leq n\mathrm{Var}(\hat{\eta}(\boldsymbol{\theta})), \quad \text{for any } \boldsymbol{\theta}\in\Theta.
\end{equation}
\end{lem}
By Taylor expansion,
\begin{equation}
\frac{p(\boldsymbol{x}| \boldsymbol{\hat{\theta}}_k)}{p(\boldsymbol{x}|\boldsymbol{ \widetilde{\theta}}_k)}=1+(\log p(\boldsymbol{x}|\boldsymbol{\hat{\theta}}_k)-\log p(\boldsymbol{x}| \boldsymbol{\widetilde{\theta}}_k))+O_p(\|\widetilde{\boldsymbol{\theta}}_k-\boldsymbol{\hat{\theta}}_k\|^2),
\end{equation}
we will have
\begin{equation*}
\widetilde{\eta}(\boldsymbol{\theta})=\sum_{k=1}^d[\frac1n\sum_{j=1}^n(1+(s(\boldsymbol{\widetilde{x}}_j^k; \boldsymbol{\hat{\theta}}_k)-s(\boldsymbol{\widetilde{x}}_j^k; \boldsymbol{\widetilde{\theta}}_k)))s(\boldsymbol{\widetilde{x}}_j^k;\boldsymbol{\theta})+O_p(\|\boldsymbol{\widetilde{\theta}}_k-\boldsymbol{\hat{\theta}}_k\|^2)],
\end{equation*}
Since $s(\boldsymbol{x}; \boldsymbol{\hat{\theta}}_k)-s(\boldsymbol{x}; \widetilde{\boldsymbol{\theta}}_k)=\text{\.{s}}(\boldsymbol{x}; \boldsymbol{\hat{\theta}}_k)(\boldsymbol{\hat{\theta}}_k-\widetilde{\boldsymbol{\theta}}_k),$ according to equation (\ref{MLE}), we have
\begin{equation*}
\widetilde{\eta}(\boldsymbol{\theta})=\hat{\eta}(\boldsymbol{\theta})-\sum_{k=1}^d\frac1n\sum_{j=1}^ns(\boldsymbol{\widetilde{x}}_j^k;\boldsymbol{\theta})\text{\.{s}}(\boldsymbol{\widetilde{x}}_j^k; \boldsymbol{\hat{\theta}}_k)(\boldsymbol{\widetilde{\theta}}_k-\boldsymbol{\hat{\theta}}^k)+O_p(\|\boldsymbol{\widetilde{\theta}}_k-\boldsymbol{\hat{\theta}}_k\|^2),
\end{equation*}
Then according to equation (\ref{MLE}), we have
\begin{equation*}
\hat{\eta}(\boldsymbol{\theta})=\widetilde{\eta}(\boldsymbol{\theta})-\sum_{k=1}^d\mathbb{E}(s(\boldsymbol{\widetilde{x}}_j^k;\boldsymbol{\theta})\text{\.{s}}(\boldsymbol{\widetilde{x}}_j^k; \boldsymbol{\hat{\theta}}_k)\mid \boldsymbol{\hat{\theta}}_k))I(\boldsymbol{\hat{\theta}}_k)^{-1}\frac{1}{n}\sum_{j=1}^n\dot{s}(\boldsymbol{\widetilde{x}}_j^k;\boldsymbol{\hat{\theta}}_k)+O_p(\frac{d}{n}),
\end{equation*}
Denote $\hat{\xi}(\boldsymbol{\theta})=-\sum_{k=1}^d\mathbb{E}(s(\boldsymbol{\widetilde{x}}_j^k;\boldsymbol{\theta})\text{\.{s}}(\boldsymbol{\widetilde{x}}_j^k; \boldsymbol{\hat{\theta}}_k)\mid\boldsymbol{\hat{\theta}}_k))I(\boldsymbol{\hat{\theta}}_k)^{-1}\frac{1}{n}\sum_{j=1}^n\dot{s}(\boldsymbol{x}_j^k;\boldsymbol{\hat{\theta}}_k)$. According to Henmi et al. (2007), $\hat{\xi}(\boldsymbol{\theta})$ is the orthogonal projection of $\hat{\eta}(\boldsymbol{\theta})$ onto the linear space spanned by the score vector component for each $\boldsymbol{\hat{\theta}}_k$, where $k\in [d]$. Then we will have
$\mathrm{Var}(\hat{\eta}(\boldsymbol{\theta}))=\mathrm{Var}(\widetilde{\eta}(\boldsymbol{\theta}))+\mathrm{Var}(\hat{\xi}(\boldsymbol{\theta})).$ Therefore, $n\mathrm{Var}(\widetilde{\eta}(\boldsymbol{\theta}))\leq n\mathrm{Var}(\hat{\eta}(\boldsymbol{\theta})).$
\begin{thm}
Under the Assumption \ref{assump}, for any $\{\boldsymbol{\hat{\theta}}_{k}\}$, we have that $$\text{as } n\to\infty,\quad n\mathbb{E}\|\boldsymbol{\hat{\theta}}_{\KL-W}-\boldsymbol{\theta}_{\KL}^*\|^2\leq n\mathbb{E}\|\boldsymbol{\hat{\theta}}_{\KL}-\boldsymbol{\theta}_{\KL}^*\|^2.$$
\end{thm}
{\bf Proof:} From Equation (\ref{KLweigthed}), we know
$$\sum_{k=1}^d\frac1n\sum_{j=1}^n\frac{p(\boldsymbol{\widetilde{x}}_j^k| \boldsymbol{\hat{\theta}}_k)}{p(\boldsymbol{\widetilde{x}}_j^k| \boldsymbol{\widetilde{\theta}}_k)}\text{\.{s}}(\boldsymbol{\widetilde{x}}_j^k;\boldsymbol{\hat{\theta}}_{\KL-W})=0.$$
Since $\frac{p(\boldsymbol{x}| \boldsymbol{\hat{\theta}}_k)}{p(\boldsymbol{x}| \boldsymbol{\widetilde{\theta}}_k)}=\exp\{\log p(\boldsymbol{x}| \boldsymbol{\hat{\theta}}_k)-\log p(\boldsymbol{x}| \boldsymbol{\widetilde{\theta}}_k)\}=1+(\log p(\boldsymbol{x}| \boldsymbol{\hat{\theta}}_k)-\log p(\boldsymbol{x}| \boldsymbol{\widetilde{\theta}}_k))+O_p(\|\boldsymbol{\widetilde{\theta}}_k-\boldsymbol{\hat{\theta}}_k\|^2),$  we have
\begin{equation}
\label{KLapprox}
\sum_{k=1}^d\frac1n\sum_{j=1}^n\text{\.{s}}(\boldsymbol{x}_j^k;\boldsymbol{\hat{\theta}}_{\KL-W})-\sum_{k=1}^d[\frac1n\sum_{j=1}^n\text{\.{s}}(\boldsymbol{x}_j^k;\boldsymbol{\hat{\theta}}_{\KL-W})
\text{\.{s}}(\boldsymbol{x}_j^k;\boldsymbol{\hat{\theta}}_{k})^T(\boldsymbol{\widetilde{\theta}}_k-\boldsymbol{\hat{\theta}}_k)+O_p(\|\boldsymbol{\widetilde{\theta}}_k-\boldsymbol{\hat{\theta}}_k\|^2)]=0.
\end{equation}
From the asymptotic property of MLE, we know
$\mathbb{E}\|\boldsymbol{\widetilde{\theta}}_k-\boldsymbol{\hat{\theta}}_k\|^2=\frac1n\mathrm{trace}(I(\boldsymbol{\hat{\theta}}_k)).$
Therefore, we know $\|\boldsymbol{\widetilde{\theta}}_k-\boldsymbol{\hat{\theta}}_k\|^2=O_p(\frac{1}{n})$ and $\sum_{k=1}^d \|\boldsymbol{\widetilde{\theta}}_k-\boldsymbol{\hat{\theta}}_k\|^2=O_p(\frac{d}{n}).$

Similar to the derivation of equation (\ref{varianceKL}), according to equation (\ref{MLE}), we have the following equation,
\begin{equation*}
\begin{split}
\boldsymbol{\hat{\theta}}_{\KL-W}-&\boldsymbol{\theta}_{\KL}^*=(\sum_{k=1}^dI(\boldsymbol{\hat{\theta}}_k,\boldsymbol{\theta}_{\KL}^*))^{-1}\sum_{k=1}^d\frac{1}{n}\sum_{j=1}^n \text{\.{s}}(\boldsymbol{\widetilde{x}}_j^k;\boldsymbol{\theta}_{\KL}^*)-\\
&(\sum_{k=1}^dI(\boldsymbol{\hat{\theta}}_k,\boldsymbol{\theta}_{\KL}^*))^{-1}\sum_{k=1}^d\mathbb{E}(\text{\.{s}}(\boldsymbol{\widetilde{x}}_j^k;\boldsymbol{\hat{\theta}}_{\KL-W})^T\text{\.{s}}(\boldsymbol{\widetilde{x}}_j^k;\boldsymbol{\hat{\theta}}_{k})
\mid\boldsymbol{\hat{\theta}}_k)\frac{1}{n}\sum_{j=1}^n\dot{s}(\boldsymbol{\widetilde{x}}_j^k;\boldsymbol{\hat{\theta}}_k)=O_p(\frac{d}{n}).
\end{split}
\end{equation*}
Then we have,
\begin{equation*}
\begin{aligned}
\boldsymbol{\hat{\theta}}_{\KL}-&\boldsymbol{\theta}_{\KL}^*=\boldsymbol{\hat{\theta}}_{\KL-W}-\boldsymbol{\theta}_{\KL}^*\\
&-(\sum_{k=1}^dI(\boldsymbol{\hat{\theta}}_k,\boldsymbol{\theta}_{\KL}^*))^{-1}\sum_{k=1}^d\mathbb{E}(\text{\.{s}}(\boldsymbol{\widetilde{x}}_j^k;\boldsymbol{\hat{\theta}}_{\KL-W})^T\text{\.{s}}(\boldsymbol{\widetilde{x}}_j^k;\boldsymbol{\hat{\theta}}_{k})
\mid\boldsymbol{\hat{\theta}}_k)\frac{1}{n}\sum_{j=1}^n\dot{s}(\boldsymbol{\widetilde{x}}_j^k;\boldsymbol{\hat{\theta}}_k)=O_p(\frac{d}{n}).
\end{aligned}
\end{equation*}
According to Henmi et al.(2007), we know the second term of above equation is the orthogonal projection of $(\boldsymbol{\hat{\theta}}_{\KL}-\boldsymbol{\theta}_{\KL}^*)$ onto the linear space spanned by the score component for each $\boldsymbol{\hat{\theta}}_k$, for $k\in [d].$
Then
$$n\mathbb{E}\|\boldsymbol{\hat{\theta}}_{\KL-W}-\boldsymbol{\theta}_{\KL}^*\|^2\leq n\mathbb{E}\|\boldsymbol{\hat{\theta}}_{\KL}-\boldsymbol{\theta}_{KL}^*\|^2.$$
We complete the proof of this theorem. $\square$
\begin{thm}
Under the Assumptions \ref{assump}, when $N> n\times d$, $\mathbb{E}\|\boldsymbol{\hat{\theta}}_{\KL-W}- \boldsymbol{\theta}_{\KL}^*\|^2=O(\frac{1}{dn^2}).$
Further, its MSE for estimating the true parameter $\vv{\theta}^*$ is
\begin{align*}
\label{globalklw}
\mathbb{E}\|\boldsymbol{\hat{\theta}}_{\KL-W}-\boldsymbol{\theta}^*\|^2
=O(N^{-1}+(dn^2)^{-1}).
\end{align*}
\end{thm}


According to Equation (\ref{KLapprox}),
\begin{equation*}
\sum_{k=1}^d\frac1n\sum_{j=1}^n\text{\.{s}}(\boldsymbol{\widetilde{x}}_j^k;\boldsymbol{\hat{\theta}}_{\KL-W})-\sum_{k=1}^d\frac1n\sum_{j=1}^n\text{\.{s}}(\boldsymbol{\widetilde{x}}_j^k;\boldsymbol{\hat{\theta}}_{\KL-W})
\text{\.{s}}(\boldsymbol{\widetilde{x}}_j^k;\boldsymbol{\hat{\theta}}_{k})^T(\boldsymbol{\widetilde{\theta}}_k-\boldsymbol{\hat{\theta}}_k)=O_p(\frac{d}{n}).
\end{equation*}

Approximating the first term of the above equation by Taylor expansion, we will get
\begin{equation}
\label{Taylorapprox}
\begin{aligned}
\sum_{k=1}^d\frac{1}{n}\sum_{j=1}^n\text{\.{s}}(\boldsymbol{\widetilde{x}}_j^k;\boldsymbol{\hat{\theta}}_{\KL-W})&=\sum_{k=1}^d[\frac{1}{n}\sum_{j=1}^n\text{\.{s}}(\boldsymbol{\widetilde{x}}_j^k;\boldsymbol{\hat{\theta}}_{k})\\
&+\sum_{k=1}^d\frac{1}{n}\sum_{j=1}^n\text{\"{s}}(\boldsymbol{\widetilde{x}}_j^k;\boldsymbol{\hat{\theta}}_{k})(\boldsymbol{\hat{\theta}}_{\KL-W}-\boldsymbol{\hat{\theta}}_k)+O_p(\|\boldsymbol{\hat{\theta}}_{\KL-W}-\boldsymbol{\hat{\theta}}_k\|^2)].
\end{aligned}
\end{equation}
Since $\|\boldsymbol{\hat{\theta}}_{\KL-W}-\boldsymbol{\hat{\theta}}_k\|^2\leq \|\boldsymbol{\hat{\theta}}_{\KL-W}-\boldsymbol{\theta}_{\KL}^*\|^2+\|\boldsymbol{\theta}_{\KL}^*-\boldsymbol{\hat{\theta}}_k\|^2$, according to equation (\ref{liuproof}), then $\|\boldsymbol{\hat{\theta}}_{\KL-W}-\boldsymbol{\hat{\theta}}_k\|^2=O_p(\|\boldsymbol{\hat{\theta}}_{\KL-W}-\boldsymbol{\theta}_{\KL}^*\|^2+\frac{d}{N}).$
We can easily derive $\text{\.{s}}(\boldsymbol{\widetilde{x}}_j^k;\boldsymbol{\hat{\theta}}_{\KL-W})=\text{\.{s}}(\boldsymbol{\widetilde{x}}_j^k;\boldsymbol{\hat{\theta}}_{k})+O_p(\boldsymbol{\hat{\theta}}_{\KL-W}-\boldsymbol{\hat{\theta}}_{k})$ for $k\in[d].$ When $N > n\times d$, we will have
\begin{equation}
\label{WeigthedFinal}
\begin{split}
\sum_{k=1}^d\frac{1}{n}\sum_{j=1}^n&\text{\.{s}}(\boldsymbol{\widetilde{x}}_j^k;\boldsymbol{\hat{\theta}}_{k})+\sum_{k=1}^d\frac{1}{n}\sum_{j=1}^n\text{\"{s}}(\boldsymbol{\widetilde{x}}_j^k;\boldsymbol{\hat{\theta}}_{k}) (\boldsymbol{\hat{\theta}}_{\KL-W}-\boldsymbol{\hat{\theta}}_k)\\
&-\sum_{k}\frac{1}{n}\sum_{j=1}^n\text{\.{s}}(\boldsymbol{x}_j^k;\boldsymbol{\hat{\theta}}_{k})\text{\.{s}}(\boldsymbol{\widetilde{x}}_j^k;\boldsymbol{\hat{\theta}}_{k})^T
(\boldsymbol{\widetilde{\theta}}_k-\boldsymbol{\hat{\theta}}_k)+O_p(\|\boldsymbol{\hat{\theta}}_{\KL-W}-\boldsymbol{\theta}_{\KL}^*\|^2)=O(\frac{d}{n}).
\end{split}
\end{equation}
$\frac{1}{n}\sum_{j=1}^n\text{\"{s}}(\boldsymbol{\widetilde{x}}_j^k;\boldsymbol{\hat{\theta}}_{k})= I(\boldsymbol{\hat{\theta}}_{k})+o_p(\frac{1}{n})$ and we also know that $\frac{1}{n}\sum_{j=1}^n\text{\.{s}}(\boldsymbol{\widetilde{x}}_j^k;\boldsymbol{\hat{\theta}}_{k})\text{\.{s}}(\boldsymbol{\widetilde{x}}_j^k;\boldsymbol{\hat{\theta}}_{k})^T=I(\boldsymbol{\hat{\theta}}_{k})+o_p(1).$ From (\ref{asym}), we know  $\boldsymbol{\theta}_{\KL}^*-\boldsymbol{\hat{\theta}}_{k}=o_p(\frac{d}{N})=o_p(\frac{1}{n}).$ When $N>n\times d,$ we have
\begin{equation}
\label{WeigthedFinal}
\begin{split}
\sum_{k=1}^d\frac{1}{n}\sum_{j=1}^n\text{\.{s}}(\boldsymbol{\widetilde{x}}_j^k;\boldsymbol{\hat{\theta}}_{k})+&\sum_{k=1}^d I(\boldsymbol{\hat{\theta}}_{k})(\boldsymbol{\hat{\theta}}_{\KL-W}-\boldsymbol{\theta}_{\KL}^*)\\
&+\sum_{k=1}^d\frac{1}{n}I(\boldsymbol{\hat{\theta}}_{k})
(\boldsymbol{\widetilde{\theta}}_k-\boldsymbol{\hat{\theta}}_k))
+O_p(\|\boldsymbol{\hat{\theta}}_{\KL-W}-\boldsymbol{\theta}_{\KL}^*\|^2)=O(\frac{d}{n}).
\end{split}
\end{equation}
Based on the Equation (\ref{MLE}), the first term and the third term of Equation (\ref{WeigthedFinal}) are cancelled. By some simple calculation, we will get
\begin{equation}
\label{Weightedorder}
n^2(\boldsymbol{\hat{\theta}}_{\KL-W}-\boldsymbol{\theta}_{\KL}^*)^T(\sum_{k=1}^dI(\boldsymbol{\hat{\theta}}_{k}))(\sum_{k=1}^d I(\boldsymbol{\hat{\theta}}_{k}))(\boldsymbol{\hat{\theta}}_{\KL-W}-\boldsymbol{\theta}_{\KL}^*)=O_p(d).
\end{equation}
This indicates, $\mathrm{Cov}(n(\sum_{k=1}^dI(\boldsymbol{\hat{\theta}}_{k}))(\boldsymbol{\hat{\theta}}_{\KL-W}-\boldsymbol{\theta}_{\KL}^*),n(\sum_{k=1}^dI(\boldsymbol{\hat{\theta}}_{k}))(\boldsymbol{\hat{\theta}}_{\KL-W}-\boldsymbol{\theta}_{\KL}^*))=O(d)$ as $n\to\infty.$ We know $n^2\mathbb{E}\|\boldsymbol{\hat{\theta}}_{\KL-W}- \boldsymbol{\theta}_{\KL}^*\|^2=\mathrm{trace}(\mathrm{Cov}(n(\boldsymbol{\hat{\theta}}_{\KL-W}-\boldsymbol{\theta}_{\KL}^*),n(\boldsymbol{\hat{\theta}}_{\KL-W}-\boldsymbol{\theta}_{\KL}^*))$. According to Assumption \ref{assump}, $I(\boldsymbol{\hat{\theta}}_{k})$ is positive definite and then $\mathrm{trace}(\sum_{k=1}^dI(\boldsymbol{\hat{\theta}}_{k}))=O(d).$ Therefore, we have
$$\mathbb{E}\|\boldsymbol{\hat{\theta}}_{\KL-W}- \boldsymbol{\theta}_{\KL}^*\|^2=O(\frac{d}{d^2n^2})=O(\frac{1}{dn^2}).$$
We know $\mathbb{E}\|\boldsymbol{\theta}_{\KL}^*-\vv \theta^*\|^2=O(N^{-1})$ from \citet{liu2014distributed}. Then we have
$$\mathbb{E}\|\boldsymbol{\hat{\theta}}_{\KL-W}-\boldsymbol{\theta}^*\|^2\approx\mathbb{E}\|\boldsymbol{\hat{\theta}}_{\KL-W}-\boldsymbol{\theta}^*_{\KL}\|^2+\mathbb{E}\|\boldsymbol{\theta}_{\KL}^*-\boldsymbol{\theta}^*\|^2=O(N^{-1}+(dn^2)^{-1}).$$
The proof of this theorem is complete. $\square$

\end{document}